%% file: main.tex
\documentclass{article}

\usepackage{microtype}
\usepackage{graphicx}
\usepackage{subfigure}
\usepackage{xcolor}
\usepackage{hyperref}

\usepackage[accepted]{icml2020}

\usepackage[utf8]{inputenc} % allow utf-8 input
\usepackage[T1]{fontenc}    % use 8-bit T1 fonts
\usepackage{url}            % simple URL typesetting
\usepackage{booktabs}       % professional-quality tables
\usepackage{amsfonts}       % blackboard math symbols
\usepackage{nicefrac}       % compact symbols for 1/2, etc.
\usepackage{color}
\usepackage{varwidth}
\usepackage{xspace}

\input{math_commands.tex}

\newcommand{\method}[0]{CARL}
\newcommand{\methodstate}[0]{\texttt{CARL (State)}}
\newcommand{\methodreward}[0]{\texttt{CARL (Reward)}}
\newcommand{\methodfull}[0]{Cautious Adaptation in RL}

\usepackage{mathtools}
\usepackage{amsmath,amssymb}
\usepackage{bm}
\usepackage{siunitx}
\usepackage{wrapfig}
\usepackage{enumitem}
\icmltitlerunning{Cautious Adaptation for Reinforcement Learning}

\begin{document}

\twocolumn[
%\icmltitle{Hope For The Best But Prepare For The Worst: \\Cautious Adaptation in RL Agents}
\icmltitle{Cautious Adaptation For Reinforcement Learning in Safety-Critical Settings }



\icmlsetsymbol{equal}{*}

\begin{icmlauthorlist}
\icmlauthor{Jesse Zhang}{Berk}
\icmlauthor{Brian Cheung}{Berk}
\icmlauthor{Chelsea Finn}{Stan}
\icmlauthor{Sergey Levine}{Berk}
\icmlauthor{Dinesh Jayaraman}{Penn}
\end{icmlauthorlist}

\icmlaffiliation{Berk}{UC Berkeley, CA, USA}
\icmlaffiliation{Stan}{Stanford, CA, USA}
\icmlaffiliation{Penn}{University of Pennsylvania, PA, USA}
\icmlcorrespondingauthor{Jesse Zhang}{jessezhang@berkeley.edu}

\icmlkeywords{Predictive Model, Variational Inference, Planning, Imitation Learning}

\vskip 0.3in
]



\printAffiliationsAndNotice{}  % leave blank if no need to mention equal contribution

\begin{abstract}
% Humans adapt quickly to new tools and environments with unknown dynamics, such as new pair of shoes, or a new tennis racquet.
% In this paper, we deal with the problem of safe adaptation: given a model trained on a variety of past experiences for a given task, can this model learn to perform that task in a new situation while avoiding catastrophic failure. This problem setting occurs frequently in real-world reinforcement learning scenarios: a vehicle adapting to drive in a new city, a robot learning to [whatever]. While learning without catastrophic failures is exceptionally difficult, prior experience can allow us to learn models that make this much easier. These models might not directly transfer to new settings in zero-shot but, if we correctly account for epistemic uncertainty, they can enable cautious adaptation that is substantially safer than na\"{i}ve adaptation and learning from scratch. [and then present main technical idea]
%, and further integrate it with a metalearning-based approach for fast system identification during adaptation.

  Reinforcement learning (RL) in real-world safety-critical target settings like urban driving is hazardous, imperiling the RL agent, other agents, and the environment. To overcome this difficulty, we propose a ``safety-critical adaptation'' task setting: an agent first trains in non-safety-critical ``source'' environments such as in a simulator, before it adapts to the target environment where failures carry heavy costs. We propose a solution approach, \method, that builds on the intuition that prior experience in diverse environments equips an agent to estimate risk, which in turn enables relative safety through risk-averse, cautious adaptation. \method\ first employs model-based RL to train a probabilistic model to capture uncertainty about transition dynamics and catastrophic states across varied source environments. Then, when exploring a new safety-critical environment with unknown dynamics, the \method\ agent plans to avoid actions that could lead to catastrophic states. In experiments on car driving, cartpole balancing, half-cheetah locomotion, and robotic object manipulation, \method\ successfully acquires cautious exploration behaviors, yielding higher rewards with fewer failures than strong RL adaptation baselines. Website at \url{https://sites.google.com/berkeley.edu/carl}.

\end{abstract}

\input{intro.tex}
\input{related.tex}
\input{approach.tex}
\input{exp.tex}

%\subsubsection*{Acknowledgments}
%Use unnumbered third level headings for the acknowledgments. All
%acknowledgments, including those to funding agencies, go at the end of the paper.

\clearpage
%\bibliography{bibref_definitions_long,references}
\bibliography{bibref_definitions_long,main}
\bibliographystyle{icml2020}
\clearpage
\appendix
\input{appendix.tex}
\end{document}

%% file: math_commands.tex
\usepackage{amsmath,amsfonts,bm}

\def\eqref#1{equation~\ref{#1}}
\def\1{\bm{1}}

\DeclareMathAlphabet{\mathsfit}{\encodingdefault}{\sfdefault}{m}{sl}
\SetMathAlphabet{\mathsfit}{bold}{\encodingdefault}{\sfdefault}{bx}{n}

%% file: intro.tex
%! TEX root=main.tex
\begin{figure}[ht]
    \centering
    \includegraphics[width=0.4\textwidth]{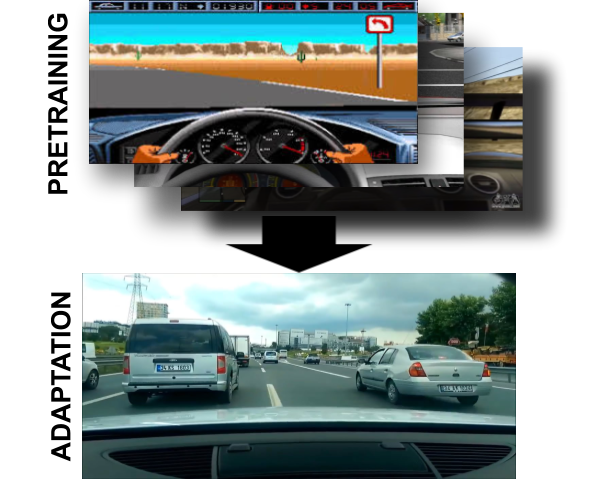}
    \caption{The Safety-Critical Adaptation (SCA) task framework. In an initial ``pretraining'' phase, an agent is trained in various sandbox source environments without safety concerns, such as in a simulator. In the second ``adaptation'' phase, it aims to adapt to a safety-critical target environment, producing high rewards quickly and safely.}
    \label{fig:carl_picture}
\end{figure}
\section{Introduction}\label{sec:intro}

%Reinforcement learning relies on iterative improvement through trial and error, and is therefore unsuitable for direct deployment in safety-critical real-world environments. An RL agent learning to drive a car on an urban street could cause great damage to its own car, as well as to the environment.
%Such problems are critical to address in autonomous systems: such as when a self-driving car must learn to drive in a new country, or when a planetary rover might have to learn to explore a harsh new environment. Missteps in real-world situations can cause real damage to robots and their environments. An important bottleneck in applying today's standard machine learning approaches to control in these real-world situations is that they are trained without any notion of safe behavior under uncertainty. Recent works have attempted to address this by proposing methods for safe exploration during reinforcement learning --- in other words, how might an agent avoid risky actions during training time? This still requires that the robot acquire its notions of uncertainty and risks at the same time as it is learning to perform tasks in the new environment, which is difficult and hazardous.

An experienced human driving a rental car for the first time is initially very aware of her lack of familiarity with the car. How sensitive is it to acceleration and braking? How does it respond to steering? How wide is the vehicle and what is its turning radius? She drives mindfully, at low speeds and making wide turns, all the while observing the car's responses and adapting to it. Within minutes, once she is familiar with the car, she begins to drive more fluently and efficiently. Humans draw upon their prior experiences to perform this kind of safe, quick adaptation to unfamiliar situations all the time, such as when playing with a new tennis racquet, or walking on a new slippery surface.

Such problems are critical to address in autonomous systems: such as when a self-driving car must learn to drive in a new country, or when a planetary rover might have to learn to explore a harsh new environment. Missteps in real-world situations can cause real damage to robots and their environments. An important bottleneck in applying today's standard machine learning approaches to control in these real-world situations is that they are trained without any notion of safe behavior under uncertainty. Recent works have attempted to address this by proposing methods for safe exploration during reinforcement learning --- in other words, how might an agent avoid risky actions during training time? This still requires that the robot acquire its notions of uncertainty and risks at the same time as it is learning to perform tasks in the new environment, which is difficult and hazardous.

Could we instead rely on transferring notions of uncertainty and risk acquired from prior experience in other related domains, such as in simulated environments, where safety may not be as much of a concern? In other words, we could make the safe learning problem easier through knowledge transfer, relaxing the problem to safe \emph{adaptation}, like the human driver above. Rather than making our planetary rover learn from scratch on a new planet, we would ask: how might it draw on its experience in varied terrains on Earth to perform cautious adaptation to the new planet? This ``safety-critical adaptation'' task framework, providing a route to real-world reinforcement learning, is the first contribution of this work.

Next, we propose a solution approach based on model-based reinforcement learning, called \methodfull\ (\method). \method\ works by first pretraining a probabilistic model of transition dynamics and catastrophe on a population of training domains with varied, unknown dynamics.
%Motivated by these questions, we propose a model-based reinforcement learning approach called risk averse domain adaptation (RADA). RADA works by first pretraining a probabilistic dynamics model on a population of training domains with varied, unknown dynamics.
Through this experience over many environments, the model learns to estimate the epistemic uncertainty (model uncertainty) of unknown environment dynamics, thus permitting estimation of a distribution of outcomes for any action executed by the agent. When introduced into a new target environment, \method\ uses this estimated distribution of outcomes to select \emph{cautious} actions that have a low chance of leading to catastrophic outcomes.
Much like the human driver in the opening example above, all the information collected during this cautious phase of exploration is fed back into the model to adapt it to the new environment, leading to increasingly well-tuned, confident predictions.
%Over time, \method\ approaches optimality in the target environment.

In our experiments, we evaluate \method\ in four safety-critical adaptation settings: driving a wider car, balancing a longer cartpole, controlling a half-cheetah robot with one disabled joint, and manipulating larger balls. \method\ consistently achieves higher rewards with fewer catastrophic failures than meta-learning and other RL baselines within few adaptation episodes in each setting. Further, \method\ successfully acquires interpretable risk-averse exploration behaviors appropriate to each setting. We view this safety-critical adaptation framework and the results of \method\ as promising initial steps towards making reinforcement learning possible in safety-critical real-world settings like robotics.

%% file: related.tex
%!TEX root=main.tex
\section{Related Work}

Cautious or risk-averse learning has close connections to learning robust control policies, as well as the uncertainty estimation derived from Bayesian reinforcement learning \citep{ghavamzadeh2015bayesian,strens2000bayesian}.
Rather than conventionally maximizing a reward function, accounting for risk usually involves allocating more attention to `worst-case' outcomes in an environment. %Such outcomes become particularly important in out-of-domain settings, where purely optimizing in the training domain does not guarantee good performance in the test domain, the problem setting that we consider in this work. %For example, a common problem in reinforcement learning is transferring a policy trained in a simulator to a real-world environment.

\noindent \textbf{Risk-Aversion in RL.~~}
Incorporating safety requires managing risks and reducing the impact of unforeseen negative outcomes. Risk management is extensively studied in quantitative finance. In portfolio optimization, a commonly used quantity that measures the expected return considering the worse $\alpha$-\% of cases is Conditional Value at Risk (CVaR) \citep{rockafellar2000optimization}. 
%With probability $\alpha$, the reward is greater than the CVaR measure. CVaR is formulated as $E[R| R \leq \upsilon_\alpha]$. 
Rather than optimizing the expected reward, risk averse policies optimize the lower $\alpha$-quartile of the distribution of rewards. 
%While meta-learning approaches like RL$^2$ \citep{duan2016rl} can potentially learn safely by adapting across learning episodes, we found this was not possible in the environments we tested. 
%To address safety more expicitly, 
To explicitly address safety, the RL community has adopted measures like CVaR into the objective \citep{morimura2010nonparametric,borkar2010risk,chow2014algorithms,tamar2015optimizing,chow2015risk}
to create risk-averse policies.
%which are robust to shifts from source to target domains. %TODO: IS THIS ONLY PLANNING, NOT RL? 
\citet{rajeswaran2016epopt} propose learning robust policies by sampling from the $\alpha$-quartile of an ensemble of models. While the model ensemble is trained on a given source distribution, the policy is only trained on the lower $\alpha$-quartile rewards from trajectories sampled from this ensemble. This leads to more conservative and robust policies.
%In contrast, we propose to train on data from varying quartiles, with the goal of preventing overly-conservative models. 

\noindent \textbf{Model-Based Safety Approaches.~~}
Prior work has utilized dynamics models to define notions of safety. One approach is to calculate safety constraints and then utilize these models, typically Gaussian Processes, to satisfy them \citep{fisac2017general, sadigh2016safe, berkenkamp2017safe, ostafew2016robust, hakobyan2015, hanssen2015, Hewing_2019}. Another approach is to incorporate the models' transition predictions into the cost function for model-free policies \citep{berkenkamp2017safe}. Other methods have used these models to perform robust model-predictive control \citep{aswani2013provably}. Generally, these works make strict assumptions, e.g. linear dynamics, quadratic cost functions, or additive state perturbations.

There are also model-based RL approaches that aim to reduce other notions of risk. \citet{yu2019modelbased} estimate future states as additional inputs to a policy maximizing financial portfolio returns and \citet{lecarpentier2019nonstationary} propose risk-averse tree search for non-stationary MDPs.

\noindent \textbf{Epistemic Uncertainty in RL.~~}
%While learning a robust model is beneficial for transferring to different domains, 
%Model-based RL offers an additional unsupervised learning signal that can be exploited at test time. In particular, 
Prior work has shown that model-based RL models can be quickly adapted during test time by meta-learning for fast adapting parameters during training \citep{nagabandi2018learning,saemundsson2018meta}. These fast-adapting parameters offer greater flexibility in adapting to new circumstances which an agent may encounter at test time. \citet{nagabandi2018learning} show that real robots can quickly adapt to miscalibrations during evaluation through this fast adaptation acquired through meta-learning. Such approaches are complementary to our approach, as they provide a means to explicitly train for fast adaptation to disturbances in the environment, while they do not account for any notion of safety.

\citet{henaff2019model} propose using the uncertainty of a model to regularize policy learning. The policy is encouraged to create trajectories which are distributionally close to trajectories observed in training data. 
In other words, the policy attempts to keep the trajectories within its training domain. In our work, our policy is encouraged to behave cautiously in unfamiliar environments rather than to explicitly remain in familiar ones. \citet{kahn2017uncertainty} train a collision prediction model in a model-based framework to favor safe, low-velocity collisions for robot navigation. 
Rather than learning to estimate uncertainties from scratch within a safety-critical environment, our approach exploits pretraining to make learning safe from the first instant in the target environment. 

\noindent \textbf{Domain randomization.~~}
Our safety-critical adaptation setting closely follows the domain randomization~\citep{sadeghi2016cad2rl,jason_sim2real2018,tobin2017domain} framework. While both attempt to pretrain policies in a mix of source environments and transfer them to some target environment, \method\ focuses on safe adaptation to the target environment --- to accomplish this, it follows an explicitly cautious action policy at adaptation time, different from the policy used in the pretraining environments.

%% file: approach.tex
%!TEX root=main.tex

%Before discussing \method, we first lay out some preliminaries.

\section{Background: PETS}\label{sec:pets}

Before discussing \method, we first lay out some preliminaries. \method\ can be built upon any model-based RL approach with ensembled dynamics models and MPC planning. For this work, we build upon PETS~\citep{chua2018deep}, a recently proposed approach for model-based reinforcement learning. We describe the main features of the PETS framework below:

\noindent\textbf{Probabilistic dynamics model.~~} PETS trains an ensemble of probabilistic dynamics models within its environment. Each model in the ensemble is a probabilistic neural network that outputs a distribution over the next state $s'$ conditioned on the current state $s$ and action $a$. The data for training these models comes from trajectories executed by following the action selection scheme described below.

\noindent\textbf{Action selection.~~} This action selection scheme is sampling-based model-predictive control (MPC): an evolutionary search method that finds action sequences with the highest predicted reward. The reward of an action sequence in turn is computed by propagating action outcomes autoregressively through the learned probabilistic models. %These predictions are produced by iterative prediction through the probabilistic models.

\noindent\textbf{Reward computation.~~} Specifically, starting from a state $s_0$, for each sampled action sequence $A=[a_1, ..., a_H]$, where $H$ is the planning horizon, the dynamics model $f$ first predicts a distribution over $s_1$ after executing $a_0$. A particle propagation method samples Monte Carlo samples from this distribution. For each sample, the dynamics model then predicts the state distribution for $s_2$, conditioned on executing $a_1$, and the process repeats. This recursive particle propagation results in a large number $N$ of particles  $\{\hat{s}^i_H\}_{i=1}^{N}$ after $H$ steps. These $N$ particles represent samples from the distribution of possible states after executing $A$. Each such particle $i\in[1,N]$ is now assigned a predicted reward $r^i$, which is a function of its full state trajectory starting from $s_0$. Finally, the mean of those predicted rewards is considered the score of the action sequence:
\begin{equation}
    R(A) = \sum_i r^i / N. \label{eq:mean_R}
\end{equation}
We call this the action score. Then, the winning action sequence $A^* = \arg\max_A R(A)$ with the highest action score is selected, the first action in that sequence is executed, and the whole process repeats starting from the resulting new state $s_1$.

%\section{FIXME: EQUATIONS}
%Original optimization objective, where $K$ is a large constant and $g(x_i)$ is an indicator of catastrophe at state $x_i$
%\begin{align*} &&\min_{x, a} \sum_{i=1}^{H}& c(x_i, a_i) \\
%&s.t. \: &x_1 &= \bar{x} \\
%&&x_{i+1} &= f_\theta(x_i, a_i) \; \forall i \in \{1...H-1\} \\
%%&&Y_i &= g_\theta (x_{i-1}, a_{i-1}) \; \forall i \in \{2...H\} \\
%&& \sum_{i=2}^H &\delta(g(x_i) = 1)  \le K
%\end{align*}
%Using the penalty method, with a hinge loss term for the inequality, we have the following equation. $\lambda$ is chosen to be a very large constant to force MPC to
%choose trajectories that do not result in catastrophic states, while also enforcing an ordering over states if the only possible trajectories include
%catastrophic states. Note that with the penalty method formulation, as $\lambda \rightarrow \infty$ the probability constraint is strictly enforced.
%\begin{align*} &&\min_{x, a} \sum_{i=1}^{H} c(x_i, a_i) &- \lambda \max\{0, \sum_{i=2}^H \delta(g(x_i) = 1) - K \}^2 \\
%&s.t. \: &x_1 &= \bar{x} \\
%&&x_{i+1} &= f_\theta(x_i, a_i) \; \forall i \in \{1...H-1\} \\
%%&&Y_i &= g_\theta (x_{i-1}, a_{i-1}) \; \forall i \in \{2...H\}
%\end{align*}
\section{Safety-Critical Adaptation with \method}\label{sec:approach}

%Now we present our approach, \methodfull\ (\method).\method\ 

We first define the ``safety-critical adaptation'' (SCA) task setting for enabling RL in safety-critical settings, as motivated in Sec~\ref{sec:intro}. SCA involves two phases:
\begin{itemize}[leftmargin=*]
    \item \textbf{Pretraining:} An agent trains in various ``source'' sandbox environments where safety is not a concern. In general, catastrophic states, even if encountered, do not cause an RL episode to terminate here.
    \item \textbf{Adaptation:} The agent must adapt to the safety-critical target environment, which generally lie outside the distribution of training environments, such as in a sim-to-real transfer setting. Catastrophic states such as car collisions, when encountered, lead to immediate RL episode termination.
\end{itemize}
We evaluate SCA approaches for their ability to achieve \emph{high task rewards, quickly, and while avoiding catastrophe}. 
%As motivated in Sec~\ref{sec:intro}, \method\ approaches safe learning as an \emph{adaptation} problem. %, where an agent may draw upon its experience in a variety of environments to guide cautious learning in a new safety-critical target environment while minimizing the risk of catastrophic failure.

To meet these requirements, we propose \method, Cautious Adaptation in RL, which builds upon PETS (Sec~\ref{sec:pets}). Although we use PETS in this work, we note CARL can be applied on most model-based RL approaches with sampling-based planners. We demonstrate this by applying \method\ to another model-based algorithm, as described in Sec~\ref{sec:exp}. We describe its behavior during the two SCA phases below.

\subsection{Pretraining: Modeling Epistemic Uncertainties}

Consider that each environment is specified by a variable $z$, which controls the state transition dynamics as well as state safety, i.e., whether a given state is safe or unsafe in that environment. In general, the target environment ID $z$ is unseen among pretraining environments.

In PETS~\cite{chua2018deep}, the probabilistic ensemble is trained to represent uncertainty within a single environment. In \method, we would instead like to train the ensemble to capture the uncertainty associated with being dropped into a new environment, with unknown environment ID $z$.

To do this, during the SCA pretraining phase, \method\ trains a single PETS agent across all pretraining environments, with varying, unknown values of $z$. Specifically, at the beginning of each pretraining episode, we randomly sample one of the pretraining $z$'s from a uniform distribution. Since $z$ 
%determines the environment dynamics and 
is unavailable to the agent,
%the uncertainty over $z$ is part of the epistemic uncertainty in this setting. Thus,
the dynamics model must capture epistemic uncertainty about state transitions and state safety due to unknown $z$. Algorithm~\ref{alg:pretraining} shows pseudocode.

%\subsection{Generalized Action Score}

%TODO

\begin{algorithm}[t]
\caption{Pretraining}
\label{alg:pretraining}
\footnotesize{
\begin{algorithmic}[1]
\STATE Initialize probabilistic ensemble dynamics model $f$ and state safety model $g$.
\STATE Collect data $\mathcal{D}$ by executing a random controller in one random training environment for one trial.
\FOR{environment ID $z\sim$ training environments}
\STATE Train the models $f$ and $g$ to predict state transitions and state safety respectively, on $\mathcal{D}$
    \FOR{$t=0 \text{ to task horizon}$}
        \FOR{evolutionary search stage=1,2,...}
            \FOR{sampled action sequence $A$}
            \STATE Run state propagation to produce $N$ particles
            \STATE Evaluate $A$ as $R(A)=\sum_i r_i/N$
            \ENDFOR
            \STATE Refine search to find $A^*=\arg\max R(A)$
      \ENDFOR
      \STATE Set $a_t$ to first action of $A^*$, and execute $a_t$.
      \STATE Record the state transition and state safety label ($w_{t+1}$) as a tuple $(s_t, a_t, s_{t+1}, w_{t+1})$  in $\mathcal{D}$
  \ENDFOR
\ENDFOR
\end{algorithmic}
}
\end{algorithm}

\subsection{Adaptation: Risk-Averse Action Selection}\label{sec:aversion}

After this pretraining, %how might the uncertainty captured in the ensemble inform cautious exploration during adaptation in the target environment?
\method\ exploits the uncertainty captured in the ensemble to perform risk-averse action selection during adaptation. We provide two versions of this scheme, representing aversion to two different notions of risk.

%To do this, we adapt the PETS action selection and reward computation scheme using a maximin notion of cautious behavior, in line with notions of risk used in prior work across disciplines~\citep{rockafellar2000optimization,tamar2015optimizing,rajeswaran2016epopt}.

\paragraph{Case 1: Low Reward Risk-Aversion.} In this case, \method\ minimizes the risk of producing low rewards over an episode. To do this, it replaces the action score $R(A)$ of~\eqref{eq:mean_R} in PETS with a newly defined ``generalized action score'' $R_\gamma(A)$, in which the ``caution parameter'' $\gamma \in [0, 100]$ controls the degree of caution exercised in evaluating action sequences in the new environment. $R_\gamma(A)$ is defined as:
\begin{equation}
    R_\gamma(A) =
        \sum_{i: r^i \leq \upsilon_{100-\gamma} ({r})} r^i/N,  \label{eq:reward_risk_averse}
\end{equation}
where $\upsilon_k ({r})$ denotes the value of the $k^{th}$ percentile of predicted rewards $\{r^j\}_{j=1}^N$ among the $N$ particles after particle propagation.

Unpacking this definition, $R_\gamma$ measures the mean score of the bottom $100-\gamma$ percentile of the predicted outcomes from the PETS model.
When $\gamma=50$, for instance, it measures the mean of the worst 50 percentile of predicted rewards. This is a pessimistic evaluation of the prospects of the action sequence $A$ --- it only pays attention to the worst performing particles in the distribution. At caution $\gamma=0$, $R_\gamma$ exactly matches the definition of $R$ in~\eqref{eq:mean_R}: it measures the mean predicted reward of all the particles produced after particle propagation through the model. In our experiments, we heuristically set $\gamma=50$.

%$R_\gamma$ measures the mean score of the top $(100-\gamma)$ percentile, i.e., all but the bottom $\gamma$ percentile, of the predicted outcomes from the PETS model. This is an optimistic evaluation of the prospects of $A$ --- it ignores the worst case possibilities and pays attention to the best case possibilities. For example, if $\gamma=50$, $R_\gamma$ is the mean of the top 50 percentile of predicted rewards.

%When caution $\gamma$ is high, $R_\gamma$ measures the mean score of the bottom $200-\gamma$ percentile of particles predicted by the probabilistic model. When $\gamma=150$, for instance, it measures the mean of the bottom 50 percentile of predicted rewards. This is a pessimistic evaluation of the prospects of the action sequence $A$ --- it only pays attention to the worst performing particles in the distribution.

To implement low reward risk-aversion, \method\ simply selects actions that optimize this generalized action score $A^{*}_\gamma=\arg\max_A R_\gamma(A)$ at adaptation time in the target environment. 

\method\ with low reward risk aversion does not require anything more than the reward function to be available at each step during pretraining. Yet, in many cases, it might automatically identify and avoid catastrophic states: for the task of driving a car to a goal, driving into a wall is catastrophic since it is highly predictive of low reward.

%\condense{Now, we define a ``$\gamma$-cautious action policy'' as one that selects actions based on the generalized action score $R_\gamma$. In other words, $A^{*}_\gamma = \arg\max_A R_\gamma(A)$. We propose to deploy such $\gamma$-cautious action policies at adaptation time.}

\paragraph{Case 2: Catastrophic State Risk-Aversion.} On the other hand, in many settings, a notion of state safety that is decoupled from the task reward is appropriate. For instance, states where the wheels of a car are on the curb may be deemed catastrophic, even if a strategy involving such states might yield high task rewards for steering around a corner.

Aversion to such catastrophic states is only possible if the pretraining environments also provide state safety annotations $w(s)$ for each state $s$ encountered by the agent. Given such annotations, during pretraining, \method\ trains an ensemble state safety predictor $g(s)$ to predict $w(s)$ alongside the ensemble dynamics models $f$ from PETS.

Then, during adaptation, rather than relying on reward-based pessimism to obtain risk-averse behavior, we directly evaluate risk by predicting the probability of encountering catastrophic states and penalize it. Specifically, rather than directly maximizing the action score $R(A)$ of Eq~\ref{eq:mean_R} during planning, we include a new penalty proportional to the estimated probability of encountering catastrophic states, as predicted by $g$.
%we would like to maximize it subject to the predicted catastrophe risk being low:
\begin{align}
  R_\lambda(A) = R(A) - \lambda g(A),
  \label{eq:cat_risk_averse}
\end{align}
where $g(A)$ is predicted probability of catastrophic failure, computed as the ensemble mean of catastrophe probabilities, summed over the predicted state trajectory. Actions are chosen to maximize $R_\lambda$. See Appendix section A for a description of how $g(A)$ is computed.

\subsection{Adaptation: Model Finetuning}

\begin{algorithm}[t]
  \caption{Adaptation}
\label{alg:adaptation}
\footnotesize{
\begin{algorithmic}[1]
\STATE \textbf{Inputs:} Pretraining dataset $\mathcal{D}$
\FOR{target environment adaptation episode=1,2,...}
    \FOR{$t=0 \text{ to task horizon}$}
        \FOR{evolutionary search stage=1,2,...}
            \FOR{sampled action sequence $A$}
            \STATE Run state propagation %to produce $N$ particles
            \STATE Evaluate $A$ with generalized score (Eq~\ref{eq:reward_risk_averse} or \ref{eq:cat_risk_averse})
            \ENDFOR
            \STATE Refine search to find $A^*=\arg\max R_\gamma(A)$
      \ENDFOR
      \STATE Execute first action of $A^*$
      \STATE Record outcome in $\mathcal{D}$
      \STATE Finetune the probabilistic ensemble model $f$ on $\mathcal{D}$
    \ENDFOR
\ENDFOR
\end{algorithmic}
}
\end{algorithm}

As it gathers experience in the target environment using its risk-averse policy, \method\ also improves its dynamics model over time to finetune it to the new environment. Since dynamics models do not need manually specified reward functions during training, the ensemble model can continue to be trained in the same way as during pretraining.

As the model improves over time, the distribution of predicted outcomes becomes more and more narrow over time. 
%For a deterministic environment, the model eventually converges to deterministic predictions, so that $R_\gamma$ is the same for all $\gamma$. In other words, once the model is well-trained, the $\gamma$-cautious action policy is identical to the standard action policy. 
The adaptation procedure in Algorithm~\ref{alg:adaptation} sums up cautious action selection and model finetuning.

Empirically, we find it useful to stabilize adaptation by maintaining data from the pretraining episodes in the training set and adding target environment data as it is acquired. \method\ computes all model updates during adaptation on this combined dataset. We use probabilistic neural network ensembles for the dynamics model~\cite{chua2018deep}, and training proceeds through stochastic gradient descent.
%This is similar to work  avoid catastrophic forgetting and stabilize adaptation using past experience.
%In our experiments, we find that training stability significantly improves through the use of this ``replay'' trick.

%Since we target fast adaptation to new environments within a handful of episodes, we propose to stabilize finetuning at adaptation time by performing model updates using a replay buffer that includes data from pretraining episodes. We propose this as a means to avoid catastrophic forgetting and stabilize adaptation using past experience. In our experiments, we find that training stability significantly improves through the use of this ``replay'' trick.

%\paragraph{Implementation details.}

%TODO TODO

%Final Algorithm Block summarizing the approach

%% file: exp.tex
%!TEX root=main.tex
\section{Experiments}
\label{sec:exp}

We evaluate \method\ for safety-critical adaptation (SCA) in four separate settings: cartpole, half-cheetah, simulated car driving, and object manipulation with a robotic hand. Recall that in SCA, the aim is to adapt effectively and quickly to a safety-critical target environment with few catastrophic events, after having trained on some sandbox source environments in the pretraining phase. 

\subsection{Experimental Setup}

\begin{figure}[t]
  \centering
  \includegraphics[height=0.15\textwidth]{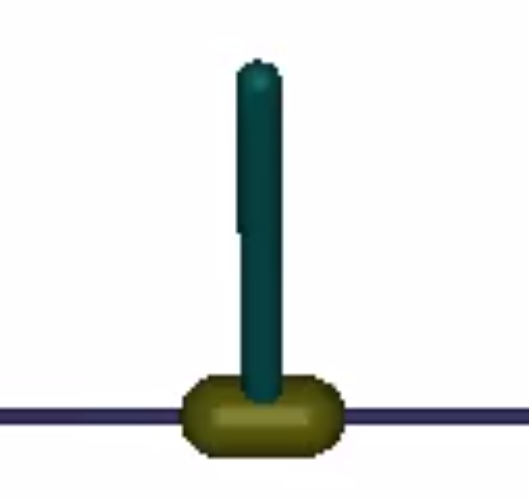}
  \includegraphics[height=0.15\textwidth]{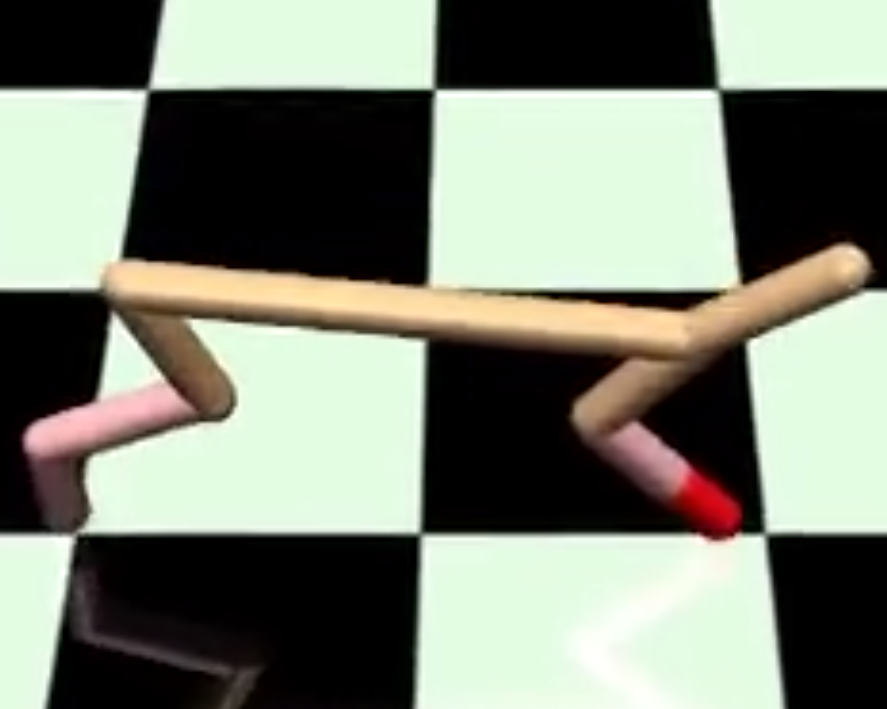}
  \includegraphics[height=0.15\textwidth]{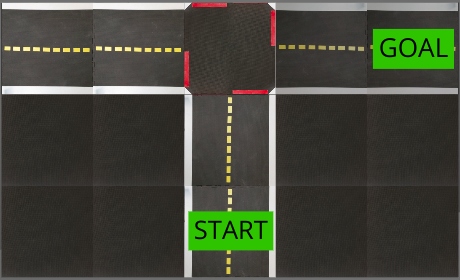}
  \includegraphics[height=0.15\textwidth]{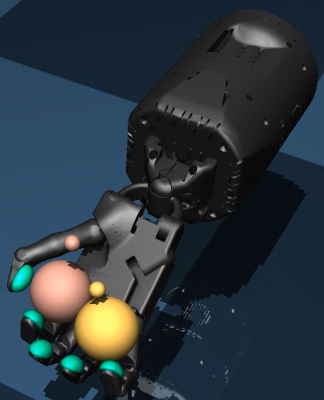}
  \caption{Images of our test environments: CartPole, Half-Cheetah Duckietown car driving, and Baoding ball manipulation.}
  \label{fig:environments}
\end{figure}

First, we describe our four safety-critical adaptation settings in detail:
\begin{itemize}[leftmargin=*]
    \item \textbf{CartPole balancing:} We modify the standard OpenAI Gym cartpole task so that the cartpole is a different length in each environment (see top left pane of Fig~\ref{fig:environments}). We randomly sample this cartpole length from the range [0.4, 0.8) during the pretraining phase.
    %the dynamics shift is caused by the length of the pole changing. 
    We define catastrophic states as those where the pole falls below the cart or the cart falls off the edge of the rail. We test on one environment from the pretraining source environment distribution (``in-distribution'') with cartpole length 0.6, and five out-of-distribution  environments, with lengths 0.3, 1.0, 1.5, 2.0, 2.5. As discussed in Sec~\ref{sec:approach}, we expect that cautious behavior is most pertinent when the target environment is significantly different from the source pretraining environments. This out-of-distribution setting also reproduces what happens when a car driving policy must be transferred sim-to-real, or when a rover is trained on Earth terrains but must work on a new planet. 
    %The pendulum length is modified as it affects how quickly the pendulum will fall catastrophically from near-catastrophic states.
    \item \textbf{Half-cheetah locomotion:} To test SCA in the Gym half-cheetah setting, %environment is identical to one presented in GrBal \citep{nagabandi2018learning} in which
    we randomly disable a joint in each training environment. This joint is marked in red in visualizations, see the top right pane of Fig~\ref{fig:environments}. During adaptation, a held-out joint is disabled. We define states where the head is in contact with the floor as catastrophic. %This event was chosen because we observed model-based agents struggling to recover from this event when the front foot is disabled.
    \item \noindent \textbf{Car driving:}  
    Our driving environment is based on Duckietown \citep{gym_duckietown}
    %\cc{, a physically accurate driving environment designed for sim-to-real transfer}. 
    The task, illustrated in the bottom left pane of Fig~\ref{fig:environments}, is to make a right turn around a corner to reach a fixed goal. %Each tile is fixed to a size of 0.585. 
    The agent observes its current $x, y$ coordinates, current velocity, and steering angle. The reward at each time step is the negative Manhattan distance from the goal, with a completion bonus of 100 if it successfully reaches the goal. The agent controls the steering angle and velocity. We do not permit the car to drive off the road, and define catastrophic states as those where the car touches the road boundaries. Cars of different widths have different catastrophic states. %The catastrophic event occurs when the car collides with the border of the road tile like this. %If the car clips the corner, it gets stuck unless the agent has learned to reverse out from the corner and try the turn again.
    Tight turns close to the corner are risky, but can yield high rewards if executed correctly.
    %The task rewards speed, so there is incentive to make risky tight turns close to the corner. %At the same time, since there is a big price to pay for it, a cautious agent must avoid hitting the corner at all costs.
     %See Fig~\ref{fig:duckietown}.
    During pretraining, the car width is sampled uniformly from the range [0.05, 0.1) before each training episode. We evaluate adaptation to widths 0.075, 0.1, 0.125, 0.15, 0.175, and 0.20.  %Recall that the width of the car is not directly provided as input to the policy, so that it cannot learn a simple width-conditioned policy, forcing it to perform experience-based adaptation at test time. This is to simulate the setting, motivated above, where the  environment ID is a high-dimensional variable, potentially involving unknown and not easily measurable properties, such as the case of the real-world car, or the planetary rover.
    \item \noindent \textbf{Hand manipulation:}  
    The robotic manipulation environment, originally presented in PDDM \cite{nagab2019deep}, involves controlling a 24-DoF hand for a Baoding ball rotation task, where the objective is to rotate two Baoding balls in the palm for a fixed time horizon without dropping either ball (see Fig~\ref{fig:environments}, bottom right). A large negative reward is obtained upon dropping either ball. More details are presented in \cite{nagab2019deep}, but we modify this environment by sampling the ball size, shared by both balls, from the range [0.026, 0.027) for each pretraining episode. We evaluate adaptation to ball sizes 0.0255, 0.0265, 0.0275, 0.0280, 0.0285, and 0.029. PETS cannot solve this task, so we build CARL upon PDDM \cite{nagab2019deep} instead. PDDM differs from PETS in various ways, such as utilizing deterministic dynamics models and a stronger MPC optimization algorithm. % that incorporates covariances across time steps and unlike PETS, PDDM's dynamics model ensemble is deterministic.
\end{itemize}

\noindent \textbf{Baselines.~~} Our first baseline is metalearning with MAML \citep{finn2017modelagnostic} applied to a PPO-trained RL agent \citep{schulman2017proximal} (\texttt{PPO-MAML}) in both cartpole and half-cheetah. Metalearning is performed in the SCA pretraining phase. In our experiments, \texttt{PPO-MAML} failed to train stably on Duckietown car driving and Baoding ball manipulation, so we omit those results. To allow \texttt{PPO-MAML} to train stably, we allowed it 1500x the number of pretraining episodes as \method. 
%We do not include the Duckietown comparison as MAML failed to consistently reach the goal during the meta-training phase. 

Our second baseline is robust adversarial reinforcement learning (\texttt{RARL})~\cite{pinto2017robust}. Since we found that this approach also required more data to train well, we compare against \texttt{RARL} at 2x and 20x the number of pretraining episodes as \method. \texttt{RARL} works by training a model-free RL agent jointly with an adversary that perturbs the actions emitted by the agent. We train the adversary to apply 2D destabilizing forces on the pendulum in cartpole and 2D forces on the torso and feet of the cheetah in half-cheetah in exactly the same way as was originally done by the authors. We train the adversary to perturb the motor torques in Duckietown. It was unable to learn to manipulate the balls in the Baoding task, so we omit \texttt{RARL} from that comparison.
%For \texttt{RARL}, we instead follow the approach from where the adversary's perturbations on the pole help it generalize to unseen  environments.

Third, we compare against a simple model-based adaptation approach, \texttt{MB + Finetune}. For this, we pretrain PETS on the same set of pretraining environments as \method\, and then adapt it in the target environment by finetuning the dynamics model, while continuing to select reward-optimal actions as in standard PETS. In the Baoding ball task, we replace PETS with PDDM \cite{nagab2019deep} as PETS fails during pretraining.
%adapt without penalizing catastrophic states in the MPC objective function ().

\begin{figure*}[t]
    \centering
    \includegraphics[width=1\textwidth]{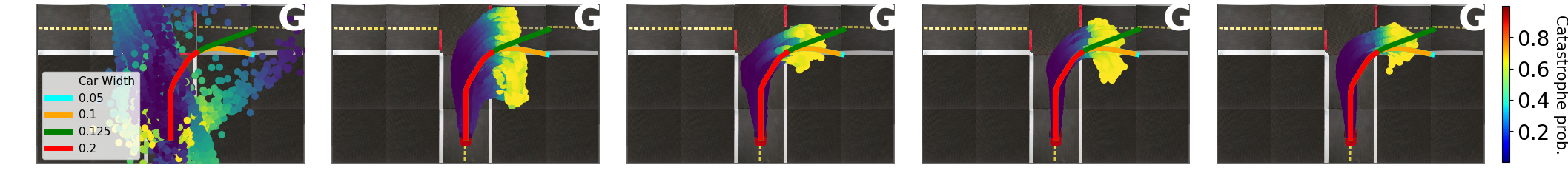}
    \caption{Trajectory and catastrophe prediction generated from a \method\ model for a fixed action sequence, at various stages of pretraining, progressing from left to right. Blue-yellow curves in the background are predicted trajectories for a fixed action sequence, and the thick curves in the foreground correspond to ground truth trajectories of different width cars when executing that action sequence. Over pretraining, the predicted trajectories converge well to the distribution of ground truth car behaviors. Predicted states are colored by the catastrophe probabilities (see color bar).}
    \label{fig:duckietown_collision_probs}
\end{figure*}
\begin{figure*}[t]
\centering
\begin{subfigure}
    \centering
    \includegraphics[width=\textwidth]{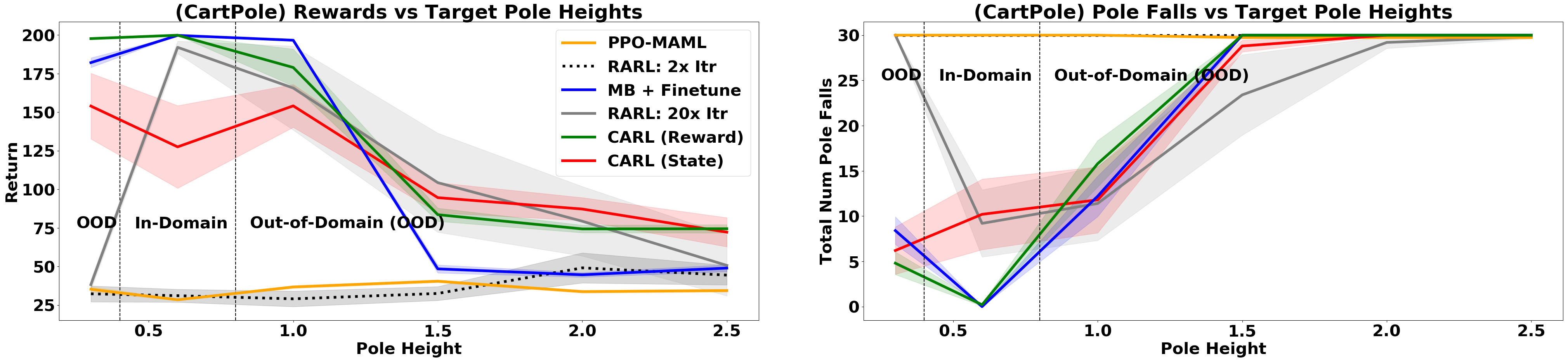}
\end{subfigure}
\begin{subfigure}
    \centering
    \includegraphics[width=\textwidth]{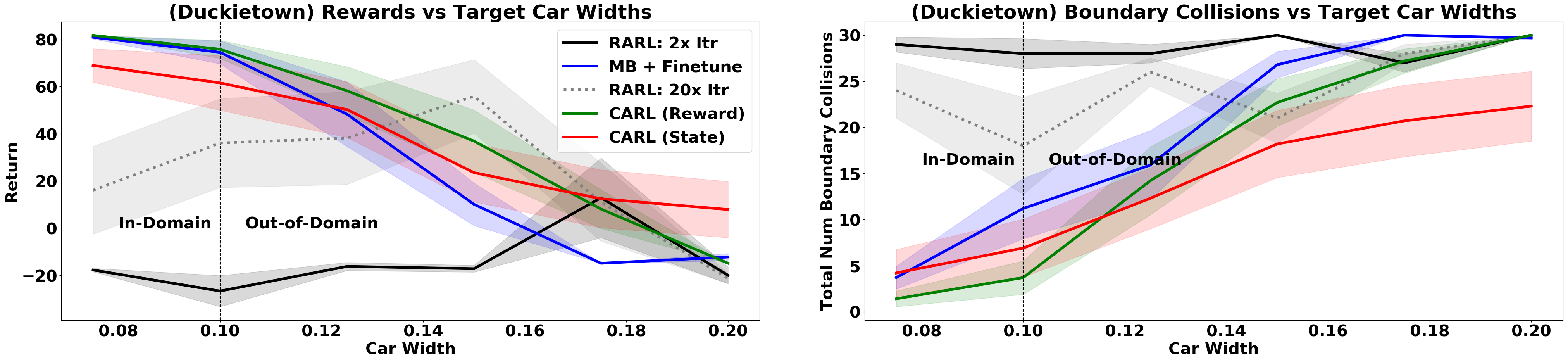}
\end{subfigure} 
\begin{subfigure}
    \centering
    \includegraphics[width=\textwidth]{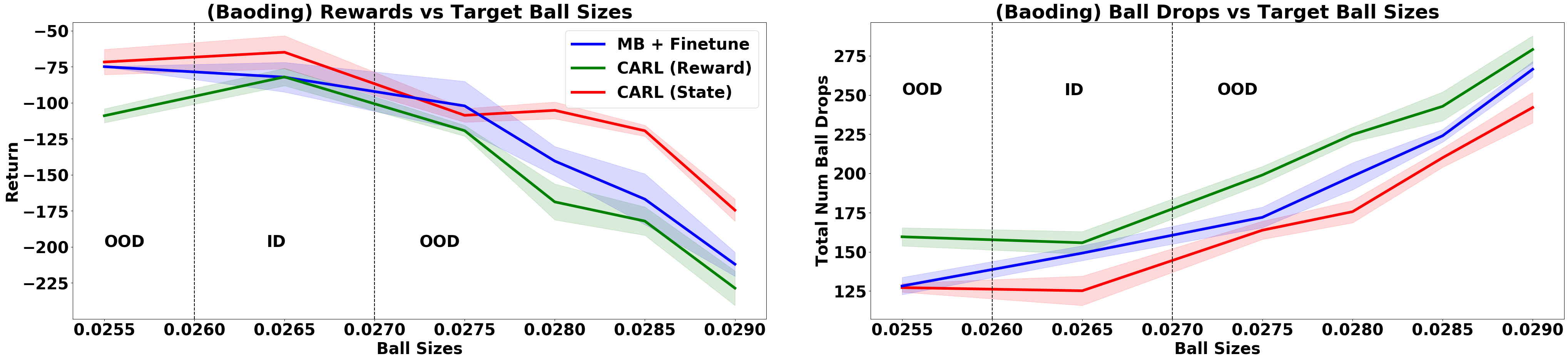}
\end{subfigure}
\caption{Evaluation of the final average maximum reward and average total catastrophic events at various target environments in the CartPole, Duckietown, and Baoding environments.}
\label{fig:duckietown_cartpole_adaptation_plots}
\end{figure*}

We also compare two variants of \method\, corresponding to the two notions of risk-aversion described in Sec~\ref{sec:aversion}. In all results, we call the low-reward risk-aversion variant as \methodreward, and the catastrophic state risk-aversion variant simply as \methodstate. \methodstate\ and \methodreward\ are built upon PDDM in the Baoding manipulation task, and upon PETS for all other environments. See Appendix section B for more experiment details.

\noindent \textbf{Performance Metrics.~~} SCA approaches aim to adapt safely to the target environment, as quickly as possible, and achieve as high rewards as possible. 
To evaluate this, for each method, we report the ``average maximum reward"  over adaptation time $t$, which is the average over 10 random seeds of the maximum over $t$ adaptation episodes of the reward obtained in the target environment. In CartPole, Duckietown, and Half-Cheetah, $t = 10$ adaptation episodes, where each adaptation episode corresponds to one environment rollout. In Baoding, $t = 15$ adaptation episodes where each episode contains $35$ environment rollouts. (Baoding policies are pre-trained on 100 episodes of $35$ environment rollouts each). Finally, to measure the safety of the adaptation process, we also report the cumulative number of catastrophes suffered by each method during adaptation, which more directly measures the extent to which different methods avoid catastrophic failures and perform safe adaptation. We keep the number of pretraining episodes fixed for all methods unless otherwise mentioned. Because SCA assumes sandbox source environments and safety-critical target environments, catastrophic events are allowed to occur at detriment to the agent's reward during the pretraining phase for our first three environments, but during adaptation these events will terminate the episode. In the Baoding environment, we also terminate episodes at pretraining time when the ball drops to avoid the discontinuous restoration of the balls to the hand.
%catastrophic events lead to resets of the dropped ball to the hand, so we terminate the episode during pretraining. % because resetting the balls after they are dropped is a sharp dynamics change that is too difficult for the models to learn.

\subsection{Results}
We now analyze the performance of \method\ and the baselines, qualitatively and quantitatively, in all four safety-critical adaptation settings.

\paragraph{Epistemic Uncertainty Visualization.} The first question we ask is: \textit{how well does the \method\ ensemble model epistemic uncertainties about state and catastrophe predictions over the course of pretraining?} To visualize this, we plot the model's predictions for the same action sequences throughout the course of training for the Duckietown environment in Fig~\ref{fig:duckietown_collision_probs}. 
On top of these trajectories predicted by the model, we overlay curves corresponding to ground truth trajectories of cars of various widths, executing the same action sequence. As pretraining proceeds in Fig~\ref{fig:duckietown_collision_probs}, the predicted trajectory distribution fits these ground truth trajectories more and more closely,
indicating that the model captures the epistemic uncertainty due to varying car widths. Furthermore, the catastrophe predictor learns the shape of the state space correctly, assigning increasing catastrophe probabilities to states as they get closer to the true catastrophe states at the road boundaries.

\paragraph{Adaptation Rewards and Catastrophes.} Given that \method\ can capture epistemic uncertainty about state transitions and catastrophic states, we now present our main quantitative results, seeking to answer, \textit{how effective is \method\ at reaching high task rewards with minimal catastrophic events during training?} 
Fig~\ref{fig:duckietown_cartpole_adaptation_plots} demonstrates the average maximum reward after all adaptation episodes and the average total number of catastrophes for each method in CartPole, Duckietown, and Baoding. 

Across CartPole and Duckietown, \methodstate\ performs slightly worse in-distribution, but attains relatively higher rewards as the testing environment moves further away from the training  environments where caution becomes important. In Duckietown, \methodstate\ generally avoids collisions better than all other methods as the car width is increased. This leads to the highest rewards relative to other methods at the largest car width, 0.2. Even in CartPole, in which nearly all methods suffered catastrophe during every adaptation step at pole heights of 1.5 and greater, \methodstate\ prevents catastrophe for longer, leading to higher rewards. Finally, in Baoding, \methodstate\ achieves higher rewards across 5 out of 6 target ball sizes than \texttt{MB + Finetune}, and drops the ball less than \texttt{MB + Finetune} on all target ball sizes. \methodreward\ performs the worst among the three methods here, matching performance with \texttt{MB + Finetune} in-distribution but obtaining lower rewards out-of-distribution and dropping the ball more across all ball sizes. However, \methodstate's ability to perform well when applied to PDDM instead of PETS demonstrates the versatility of our catastrophic state risk aversion approach.

\texttt{MB + Finetune} achieves high reward with few catastrophes on in-distribution environments, however its reward drops off drastically at out-of-distribution CartPole lengths. Its rewards also decrease while collisions increase in Duckietown as the car width increases. \texttt{RARL: 2x Itr} generally performs poorly in CartPole and Duckietown. \texttt{RARL: 20x Itr}'s performance is competitive with both \method\ variants, however it is still outperformed at the environments farthest away from the training distribution despite being pretrained on 20x the number of trajectories. \texttt{PPO-MAML} is omitted in our Duckietown results, and both \texttt{PPO-MAML} and \texttt{RARL} are omitted in our Baoding results, because they did not train stably.
On CartPole, \texttt{PPO-MAML} performs poorly, achieving low reward and reaching catastrophic states nearly every episode across in-distribution and out-of-distribution environments, indicating that it may require many more adaptation episodes. % than the rest of the methods in order to adapt to any target environment.

\begin{figure}[!htb]
	\centering
    \includegraphics[width=0.225\textwidth, trim={0 15 0 0}, clip]{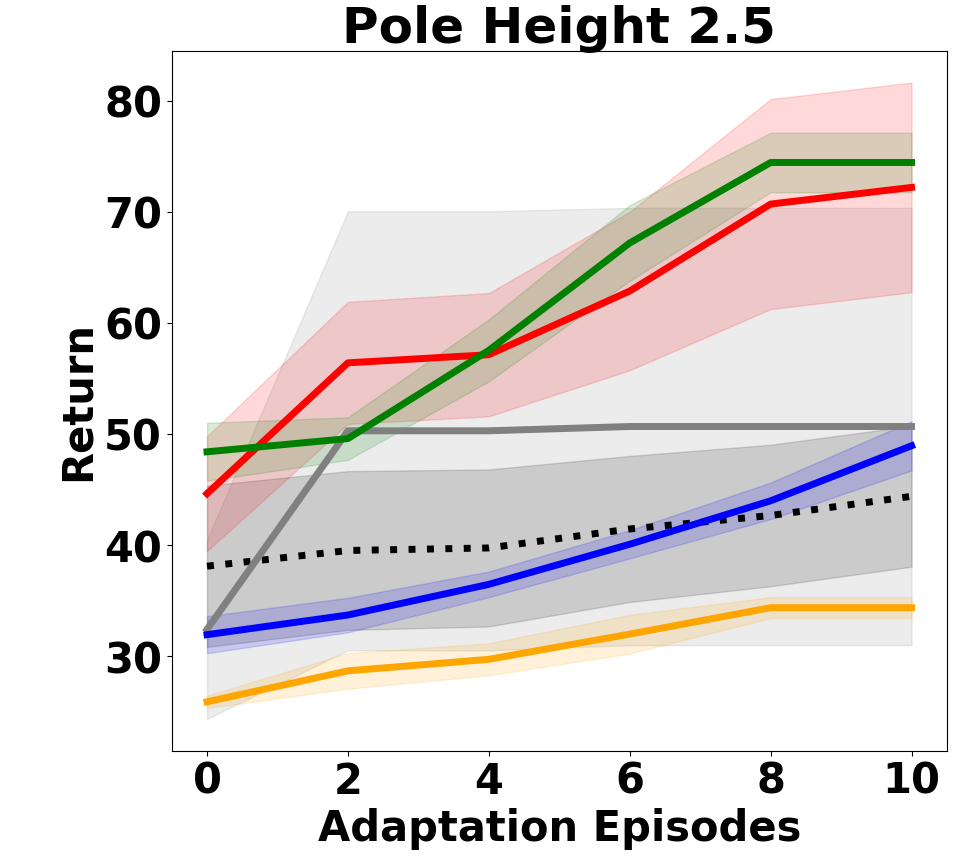}
    \includegraphics[width=0.225\textwidth, trim={0 15 0 0}, clip]{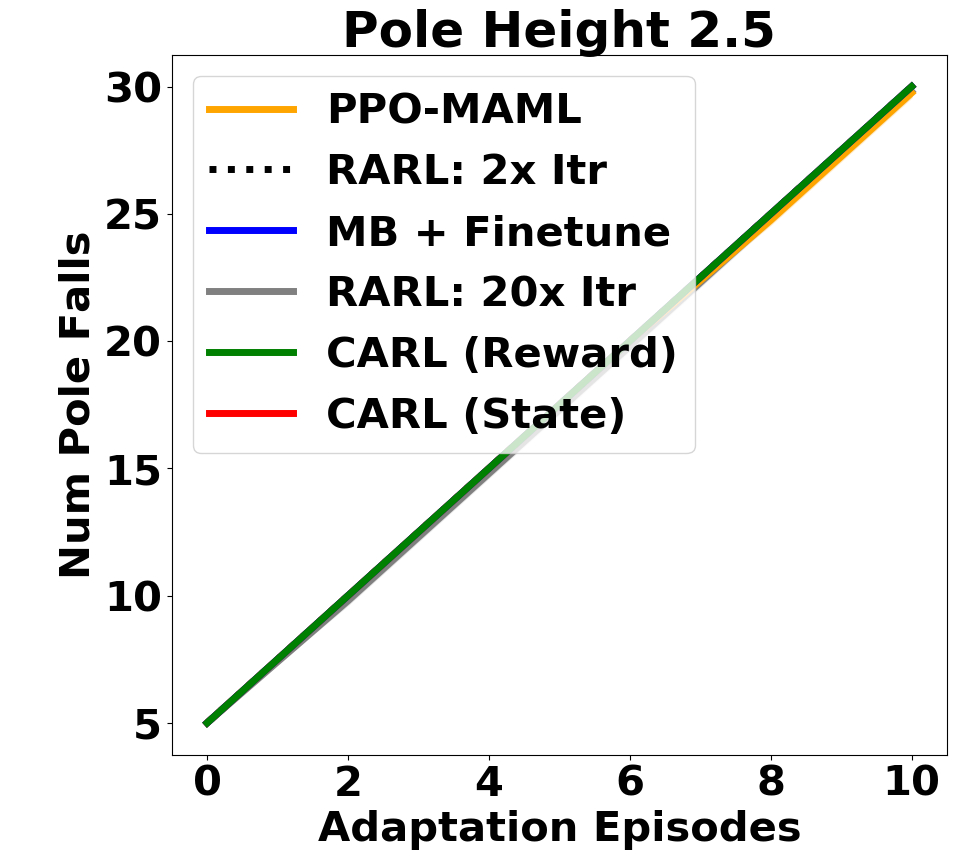}
    \includegraphics[width=0.225\textwidth, trim={0 15 0 0}, clip]{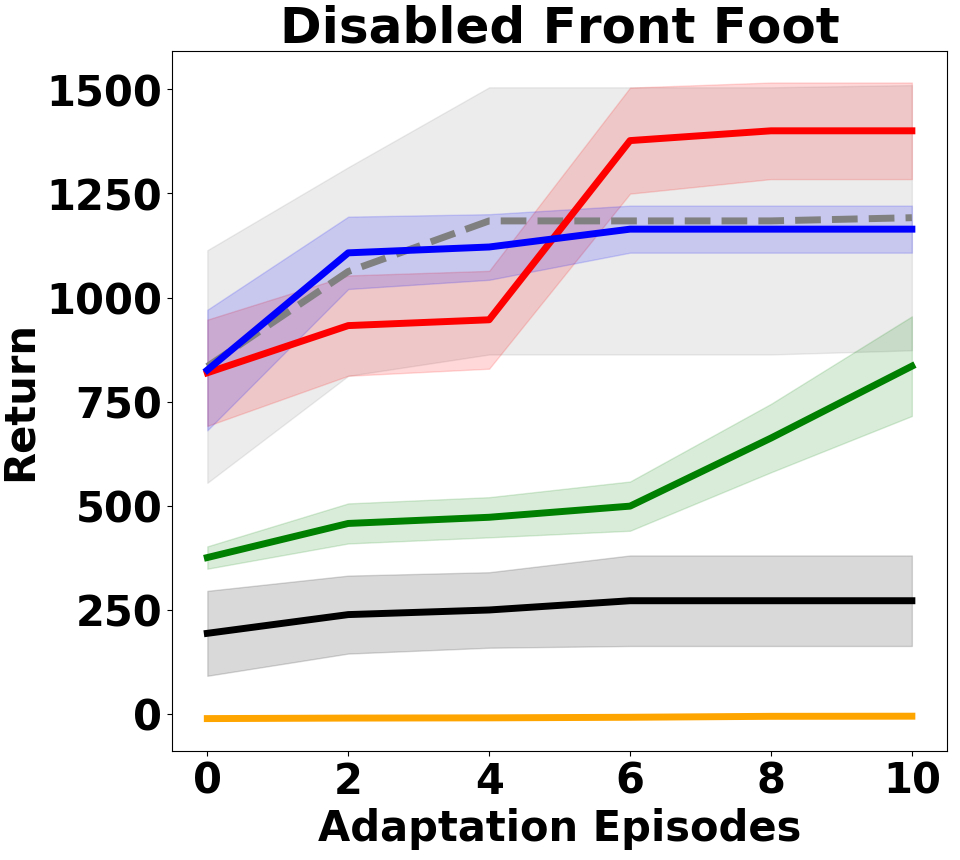}
    \includegraphics[width=0.225\textwidth, trim={0 15 0 0}, clip]{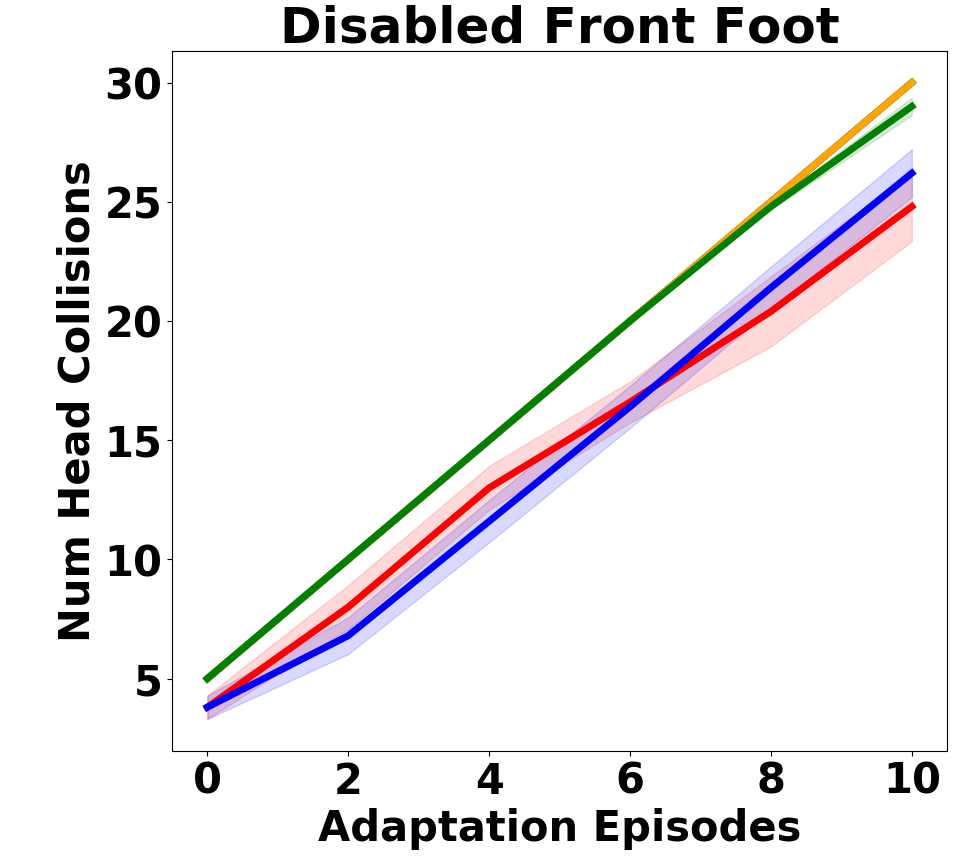}
    \includegraphics[width=0.225\textwidth]{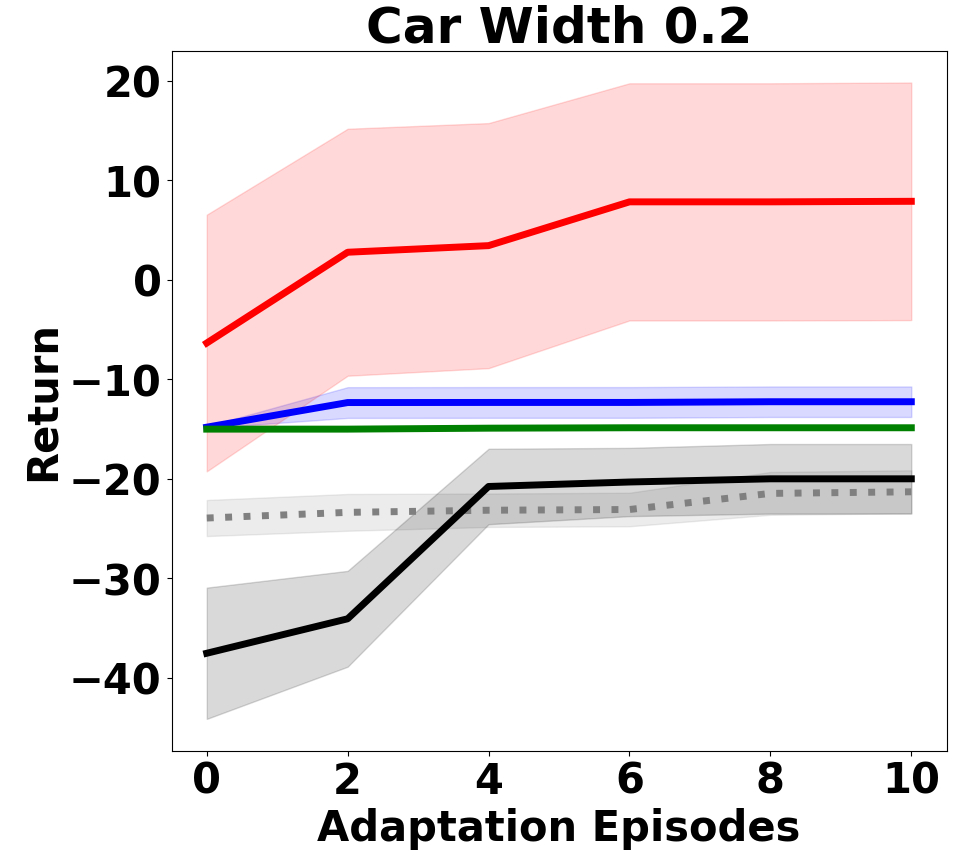}
    \includegraphics[width=0.225\textwidth]{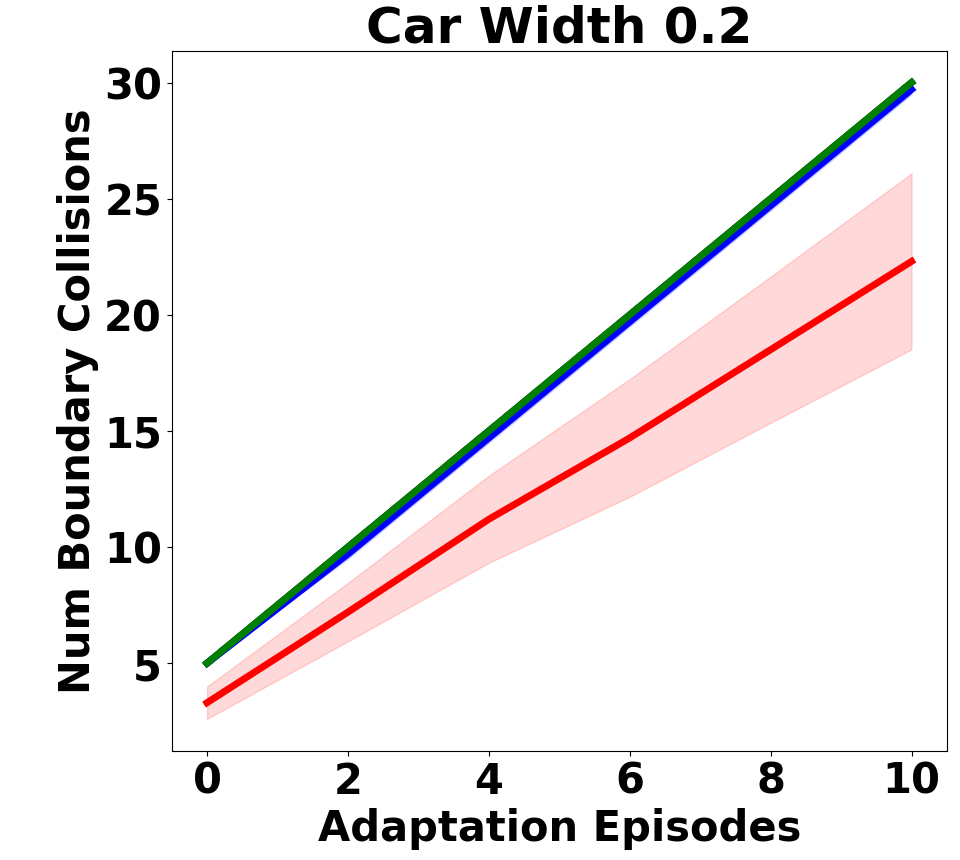}
    \includegraphics[width=0.225\textwidth]{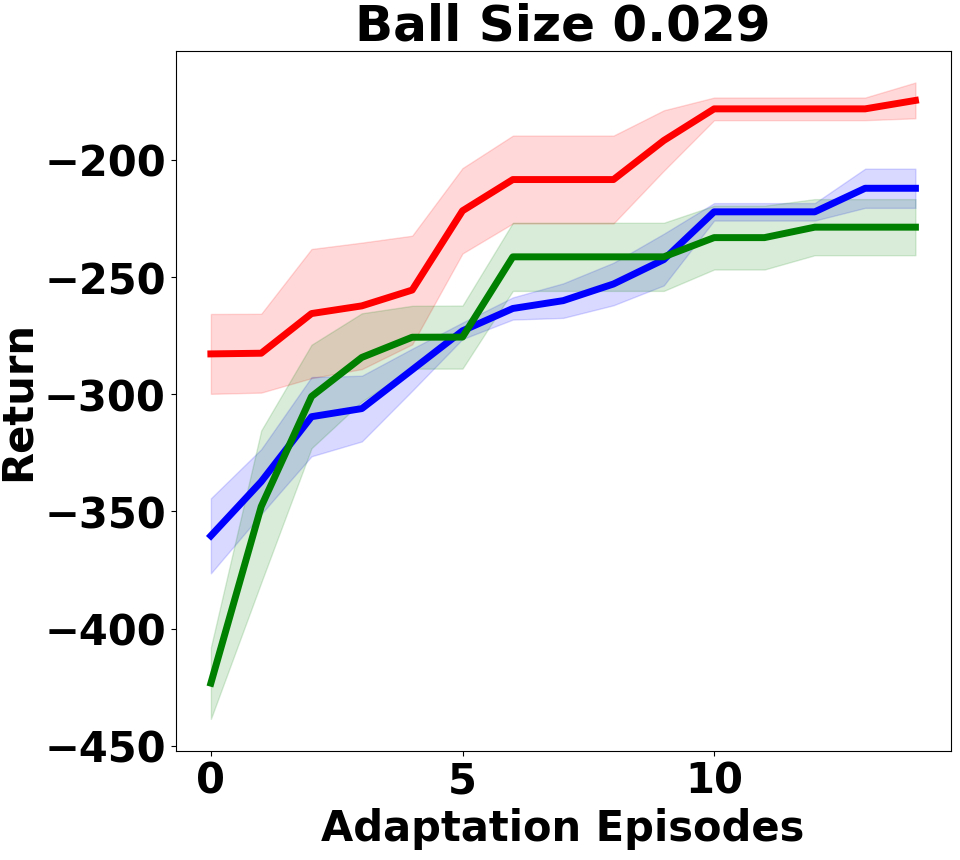}
    \includegraphics[width=0.225\textwidth]{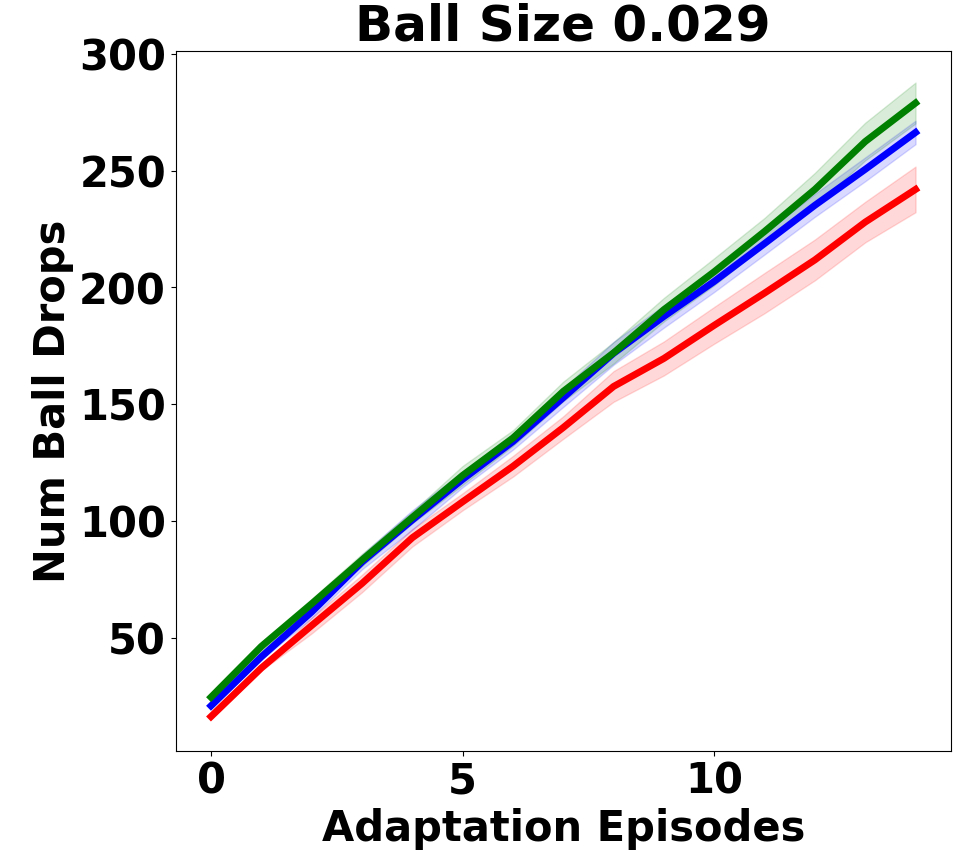}
    \caption{Average maximum reward and average total catastrophes over time during adaptation. \textit{Top:} CartPole at the largest pole height, 2.5. \textit{Second:} Half-Cheetah with the front foot disabled. \textit{Third:} Duckietown at the largest car width, 0.2. \textit{Bottom:} Baoding at the largest ball size, 0.029.}
    \label{fig:adaptation_speed_selected_plots}
\end{figure}

In Half-Cheetah, our experiments evaluate only one out-of-distribution setting: adapting to running with a disabled front foot. We report the results after 10 adaptation episodes here (also see Fig~\ref{fig:adaptation_speed_selected_plots}): \methodstate\ performs the best by obtaining rewards around 1400 and reaching the least catastrophic states across all methods. \texttt{MB + Finetune} and \texttt{RARL: 20x Itr} reach rewards nearing 1200, however \texttt{RARL: 20x} enounters catastrophe every iteration. \texttt{CARL (Reward)} reaches rewards of 750, while \texttt{RARL: 2x Itr} and \texttt{PPO-MAML} perform poorly. Most of these results match CartPole and Duckietown, with the exception of the much poorer performance of \methodreward. We see that in Half-Cheetah, reward-based risk aversion is insufficient to completely prevent reaching low-value states such as the head colliding with the ground, and state-based aversion is required.

\begin{figure}[!htb]
\centering
\begin{subfigure}
    \centering
    \includegraphics[width=0.475\textwidth]{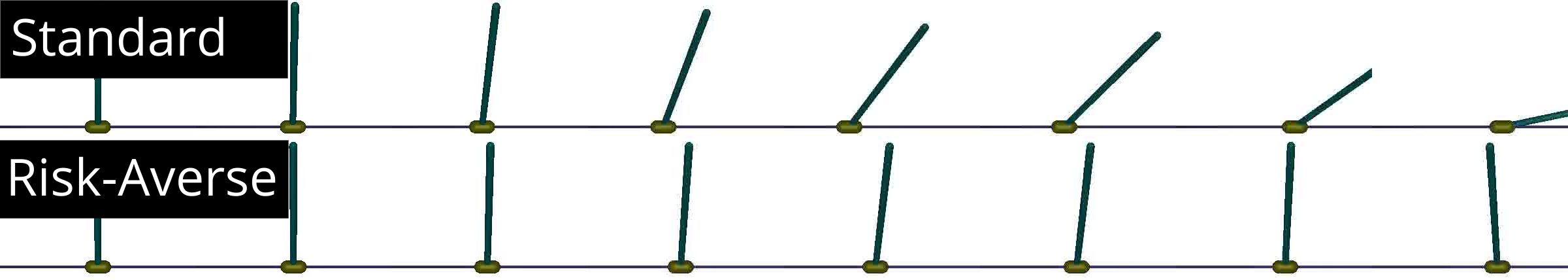}
\end{subfigure}
\begin{subfigure}
    \centering
    \includegraphics[width=0.475\textwidth]{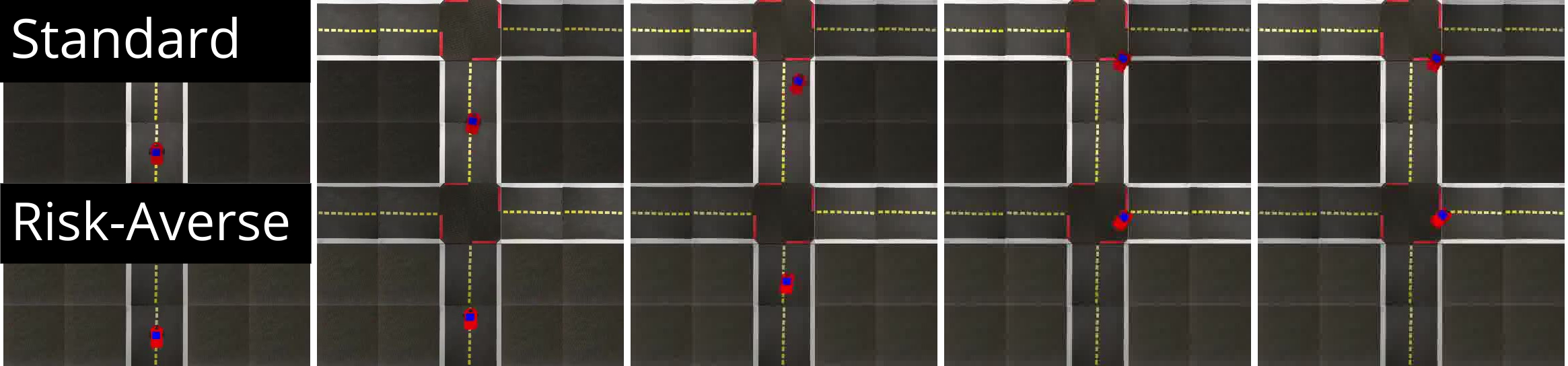}
\end{subfigure} 
\begin{subfigure}
    \centering
    \includegraphics[width=0.475\textwidth]{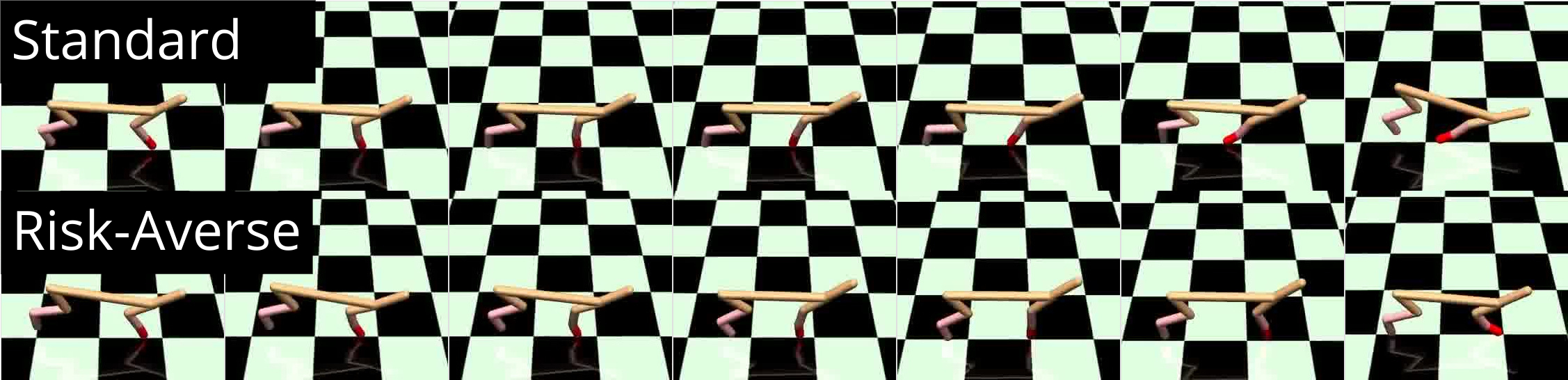}
\end{subfigure} 
\caption{A comparison of standard vs. risk-averse behavior, under the same out-of-distribution conditions and with the same dynamics model weights in three settings. In each setting, the top row contains frames from an episode of standard model-based planning (arranged left-to-right) while the bottom row shows \method\ 's catastrophic state risk-averse planning. The specific environments are: \textit{(Top)} CartPole with pole length 1.0. \textit{Middle:} Duckietown with car width 0.15. \textit{Bottom:} Half-Cheetah with the front foot disabled.}
\label{fig:rollout_reels}
\end{figure}

\paragraph{Adaptation speed.} Next, we ask, \textit{how does cautious action selection help with adaptation speed}? We plot results over adaptation time in each environment for both the average maximum reward and the running total catastrophe count to show adaptation speed in the target environments. Fig~\ref{fig:adaptation_speed_selected_plots} shows the average maximum reward over adaptation time for all methods in the farthest out-of-distribution environments of CartPole, Duckietown, and Baoding. It also demonstrates the reward over time on the one target environment of Half-Cheetah. In CartPole (top row), risk-aversion allows both \method\ methods to immediately start with higher returns on the first iteration. As adaptation progresses their lead over other methods continues and it results in both achieving max returns above 70. All other methods are unable to significantly improve within 10 adaptation steps. Even though all methods experience catastrophe at nearly every iteration, the \method\ approaches both obtain higher returns by delaying catastrophe.

For the Half-Cheetah task (second row), \texttt{RARL: 20x Itr}, \texttt{MB + Finetune}, and \methodstate\ obtain similar rewards upon starting. However, halfway through adaptation, \methodstate\ achieves and maintains the best performance afterwards. \methodreward\ starts off with much lower rewards, but eventually improves significantly. \texttt{MB + Finetune} and \texttt{RARL: 20x Itr} are also able to improve, however \texttt{RARL: 2x Itr} and \texttt{PPO-MAML} fail to adapt to the target environment. 

\methodstate\ performs much better than the other methods in Duckietown (third row). Catastrophic state risk-aversion allows it to achieve higher returns from the start, and continue to adapt on the test environment. While all other methods collide with the border nearly every iteration, \methodstate\ collides significantly less, and even reduces the collision rate throughout the course of adaptation. Interestingly, \texttt{RARL: 2x Itr} is also able to increase its rewards during adaptation, whereas every other method fails to significantly improve its performance. We note that in the driving task, positive rewards are received only upon reaching the goal. Thus \methodstate\ is the only method able to successfully make the turn in this extremely out-of-distribution task.

Finally, in the Baoding task, \methodstate\ is able to achieve substantially higher reward from the first adaptation episode, and it continues to adapt and achieve higher reward than both other methods throughout the course of adaptation. This is mirrored in the catastrophe plot, in which \methodstate\ consistently drops the ball the least number of times during adaptation. Here, \methodreward\ and \texttt{MB + Finetune} perform similarly.

\paragraph{Cautious Behavior} Finally, given the benefits of risk-aversion, we ask, \textit{what qualitative behaviors does risk aversion produce}? To answer this, Fig~\ref{fig:rollout_reels} presents state trajectories produced using planning with the same dynamics model, with and without the risk aversion of \methodstate. We note that for all visualizations, our dynamics models' weights are exactly the same and we perform a frame-by-frame comparison starting from the same timestep in the rollouts; the only difference is the risk-aversion. In three environments, we see that \methodstate\ produces distinct, interpretably risk-averse behaviors. 
In CartPole, \methodstate\ is more willing to pay action costs in order to keep the pendulum upright. This behavior likely comes from how a lower pole can lead to states with higher catastrophe probability. When viewing rollouts, we notice that this behavior results in the cart using much more of the rail to do so, trading off the risk of the pole falling for the risk of falling off the rail. 
In Duckietown, the risk-averse agent drives more slowly and takes a wider turn, resulting in it being able to fully complete the turn while the standard agent clips the corner. We even notice risk-averse behaviors such as making an extremely wide turn, then backing up and completing the turn to reach the goal. 
The Half-Cheetah comparison highlights a key behavior that allows \methodstate\ to stay upright for longer. When the disabled front foot is too far back, it is still able to recover from the precarious situation by angling the front leg joints differently.

\section{Discussion}
To address the difficulty of applying reinforcement learning in safety-critical situations such as robotics, we propose a general safety-critical adaptation task setting that transfers knowledge gained in less safety-critical source environments to enable safe learning in the target environment. Further, we have proposed \method, a new method for model-based reinforcement learning approaches that works by implementing risk-averse adaptation strategies within the safety-critical adaptation framework. Our experiments demonstrate that \method\ consistently outperforms strong adaptation baselines including metalearning and robust RL on four control tasks, including a difficult object manipulation task where \method\ is applied to a new model-based RL algorithm. \method\ achieves higher rewards, more quickly, while suffering fewer catastrophic events during learning.

%% file: appendix.tex
\section{\method\ Catastrophe Probability Prediction} \label{appendix:math}
\paragraph{Constraints and Formulation}
Our planner starts from a state $s_0$, a sampled action sequence $A = [ a_1, ..., a_H]$ over a planning horizon $H$, our ensemble parameters $\{\theta_1, ..., \theta_{|E|}\}$, a caution tuning parameter $\beta$, a function $f$ parameterized by $\theta$ that maps states and actions to distributions over the next state, and a function $c$ parameterized by $\theta$ that produces collision probabilities.

We denote the predicted catastrophe cost as $g(A)$, which is the predicted sum of failure probabilities for all states along the trajectory produced by $A$. Similarly to prior work, we define failure probabilities as the probability that the state is in a set of catastrophic states, $Catastrophic Set$, for the given environment. Note that because $f$ outputs distributions over the next state, the states $s_i$ themselves are random variables.

$g(A)$ can then be decomposed into $$g(A) = \sum_{i=1}^{H} P(s_i \in CatastrophicSet)$$
where $$P(s_i \in CatastrophicSet) = \frac{\sum_{j=1}^{|E|}\delta(c_{\theta_j}(s_{i-1}, a_{i-1}) > \beta)}{|E|}$$ and $$s_i = f(s_{i-1}, a_{i-1})$$ That is, the probability of catastrophe is given by the proportion of models in the ensemble that predict a catastrophe to occur with probability greater than $\beta$. Here, we are assuming that the true probability of catastrophe can be approximated by the empirical bootstrap distribution produced by our dynamics models. The models output isotropic Gaussian distributions over the next state and the probability of catastrophe given the previous $(s_{i-1}, a_{i-1})$ pair. In practice, we do not need to divide by the ensemble size $|E|$, as the $\lambda$ coefficient on $g(A)$ in Eq. \ref{eq:cat_risk_averse} can be scaled appropriately to absorb the coefficient.

\paragraph{Connection to Constrained MDPs}
We can formulate the problem as a chance constrained MDP, where we set a bound on the probability that one or more states is in the catastrophic set. This corresponds to the constraint $\sum_{i} P(s_i \in CatastrophicSet) <= K$ for all planned states $i$ with a large constant $K$. The Lagrangian relaxation of this is the exact problem that we optimize for in MPC planning: maximizing Eq. \ref{eq:cat_risk_averse}.
%This entire process is approximated by the trajectory sampling utilized by PETS, however with a deterministic model this would be exact.

\paragraph{Training} For all of our experiments, we heuristically select $\beta$ to be $0.5$. Furthermore, we set $\lambda$ used to scale $g(A)$ in $R_{\lambda}(A)$ to be an arbitrarily high penalty value: $10000\cdot|E|$. Empirically, in order to train $c$, we simply extend the state space of our environments with a catastrophe indicator and train the catastrophe output of our dynamics models with the binary cross entropy loss function. Therefore our loss function is the addition of the cross entropy loss term with standard supervised probabilistic state loss term from PETS \cite{chua2018deep}.

\paragraph{Ablation on $\beta$} We perform an ablation on $\beta$ in appendix Fig~\ref{fig:ablation}. We see that across CartPole, Duckietown, and Half-Cheetah, setting $\beta = 0.5$ produces the best overall results. Interestingly, setting $\beta$ lower doesn't necessarily result in fewer catastrophes. All three levels of $\beta$ perform similarly in the simpler CartPole environment and the challenging Baoding environment.

However in Duckietown we see that $\beta = 0.25$ performs significantly worse than the other two. We notice that this behavior comes from the agent often refusing to make the turn at all, being overly risk-averse about making any action that could lead to hitting the right road tile boundaries. This occasionally leads to the agent then hitting other road boundaries that were not encountered during training time. And as expected, $\beta=0.75$ experiences the largest number of boundary collisions in out of domain settings as the car width increases. Note that for both CartPole and Duckietown, $\beta=0.75$ performs the best in-domain, which is to be expected as there is less need to be risk-averse in environments it was trained on. 

In Half-Cheetah, $\beta=0.25$ does allow the agent to start with a higher reward than the other two $\beta$ ablations, despite encountering a similar number of head collisions. This means the most risk-averse ablation prevents head collisions for longer than the other two methods in the first few stages of adaptation. However the best final performance is achieved by $\beta=0.5, 0.75$.

In Baoding, performance is generally similar across all three $\beta$ values. 
\section{\method\ Experiment Details}
 All model-based methods (\method\, \texttt{CARL (Reward)}, \texttt{MB + Finetune})) are trained with the same hyperparameters, which are listed in Table~\ref{tab:hyperparams}. For CartPole and Half-Cheetah, the params are chosen with little modification to the original training parameters of the similar environments in PETS \cite{chua2018deep}. And for Baoding, the parameters were chosen to reproduce the PDDM \cite{nagab2019deep} pretraining results as closely as possible with fixed ball weights, and are based on the ones listed in the paper. For all environments, the number of training iterations is relatively high to ensure the method is able to successfully solve the task and ensure \method\ has sufficient training iterations to accurately predict catastrophe probabilities. We find that training for longer produces even better pretraining and adaptation rewards for the Baoding task, however the wall-clock time for both pretraining and adaptation greatly increases. The hyperparameters have not been extensively tuned and it is likely that the parameters and loss function weighting can be adjusted to achieve better adaptation performance with fewer training iterations. We note that the model free methods we compare against are trained on many more iterations.

 We train \texttt{RARL} on both 2x and 20x the number of iterations we train the model-based methods on for a fair comparison, and we train \texttt{PPO-MAML} on 1500x the number of iterations to ensure good meta-training performance before testing adaptation.
\begin{table*}[h]
    \centering
    \begin{tabular}{c|c c c c c}
    Param/Env & CartPole & Half-Cheetah & Duckietown & Baoding \\
    \hline
    Ensemble Size & 5 & 5 & 5 & 3 \\
    \# Hidden Layers & 4 & 5 & 4 & 2 \\
    Hidden Layer Size & 500 & 200 & 200 & 500 \\
    Optimizer &  Adam & Adam & Adam & Adam \\
    Learning Rate & 1e-3 & 1e-3 & 1e-3 & 1e-3 \\
    Batch Size & 256 & 256 & 256 & 512 \\
    \# Random Itrs & 1 & 1 & 2 & 100 initial, then 5 per itr \\
    \# On-Policy Rollouts & 50 & 100 & 100 & 3000 (30 per itr, 100 itrs) \\
    Epochs per Itr & 5 & 10 & 5 & 40 \\
    Planning Horizon & 25 & 10 & 25 & 8 \\
    CEM Popsize & 400 & 500 & 400 & - \\
    CEM \# Elites & 40 & 50 & 40 & - \\
    $\beta$ (\texttt{CARL} and \texttt{CARL (Reward)}) & 0.5 & 0.5 & 0.5 & 0.5 \\
    \end{tabular}
    \caption{Table of parameters for every environment across all model-based methods. See \cite{nagab2019deep} or our code for Baoding optimizer parameters.}
    \label{tab:hyperparams}
\end{table*}
\begin{figure*}[t]
\centering
   \includegraphics[width=0.475\textwidth]{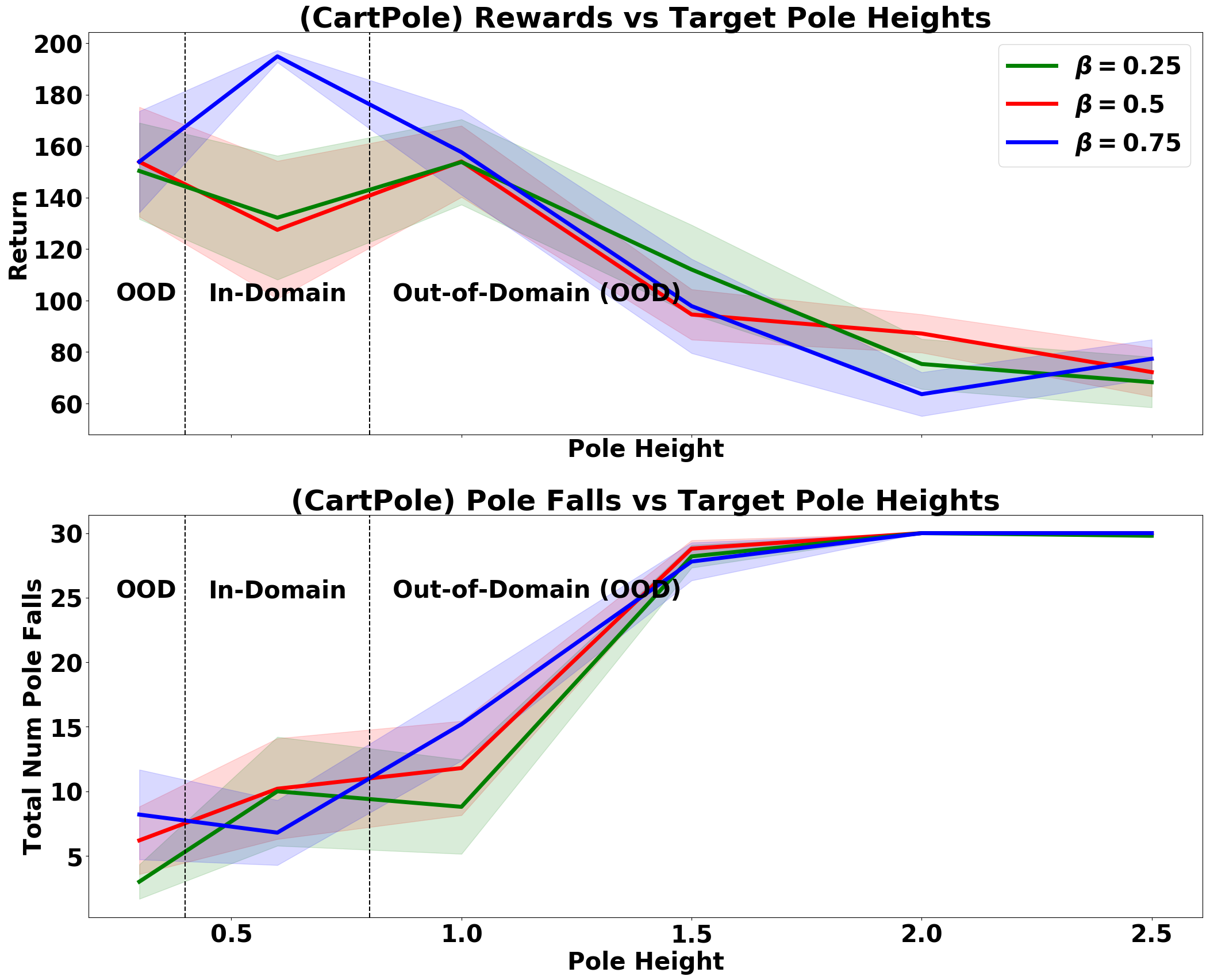}
	\includegraphics[width=0.475\textwidth]{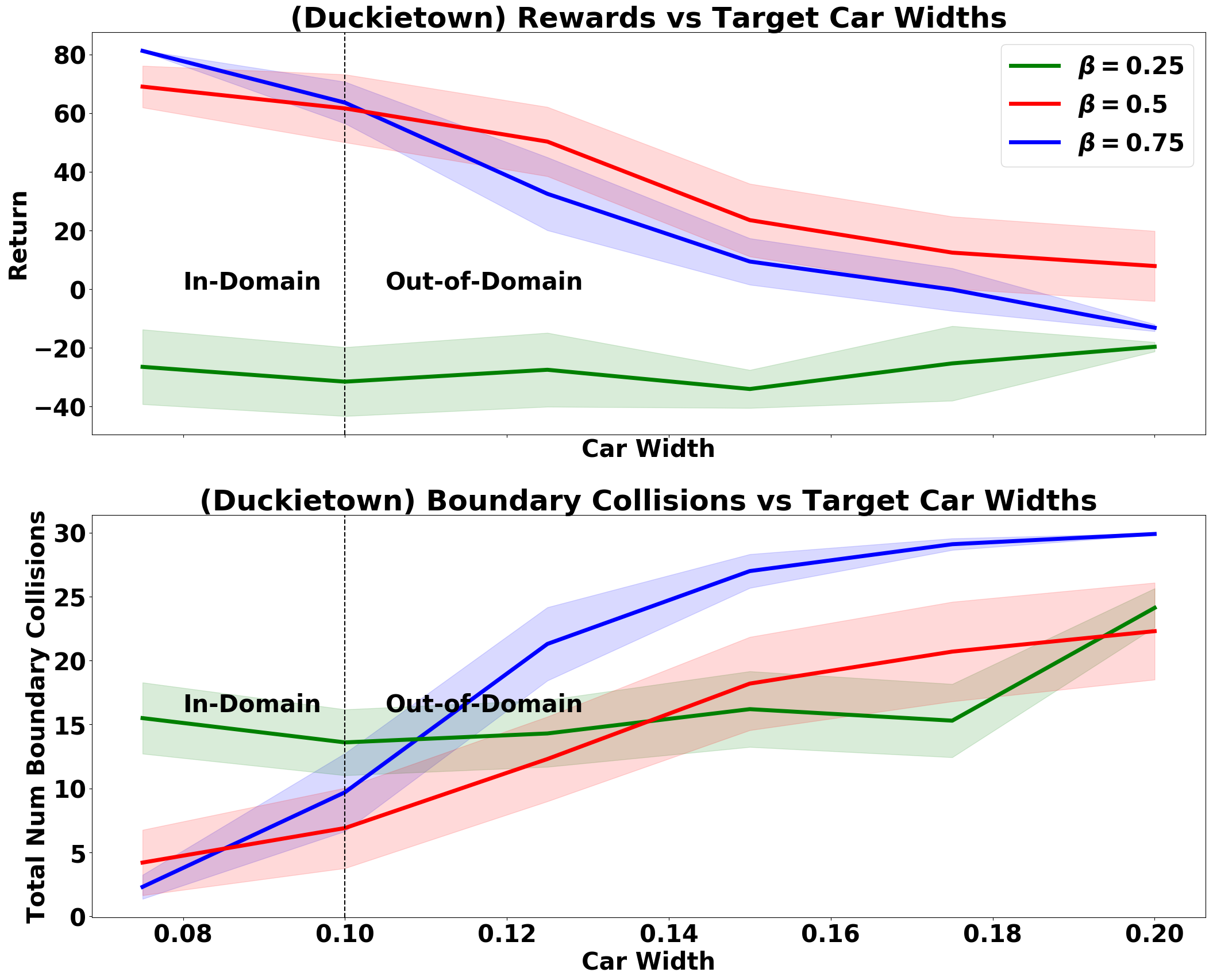}
	\includegraphics[width=0.475\textwidth]{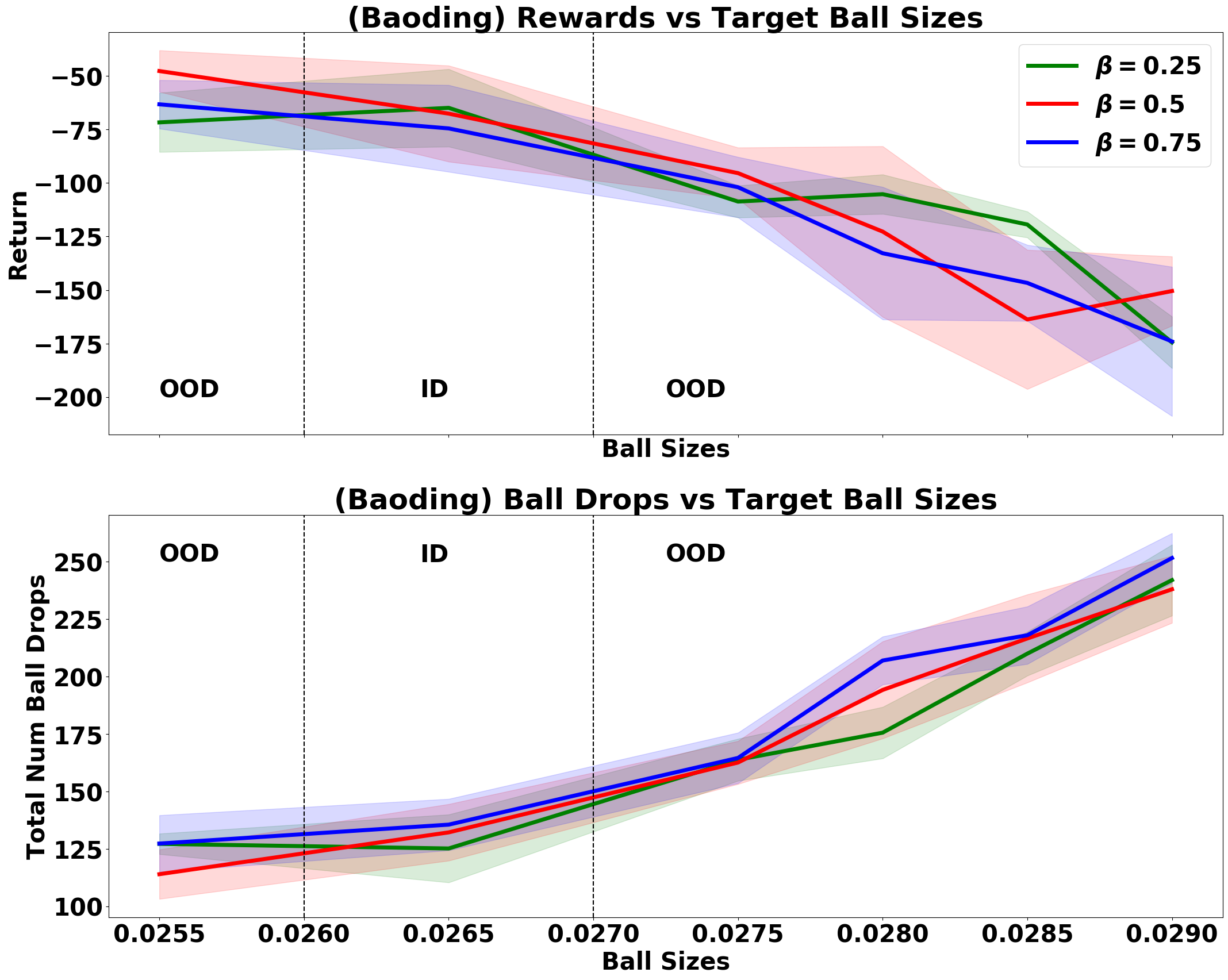}
	\includegraphics[width=0.475\textwidth]{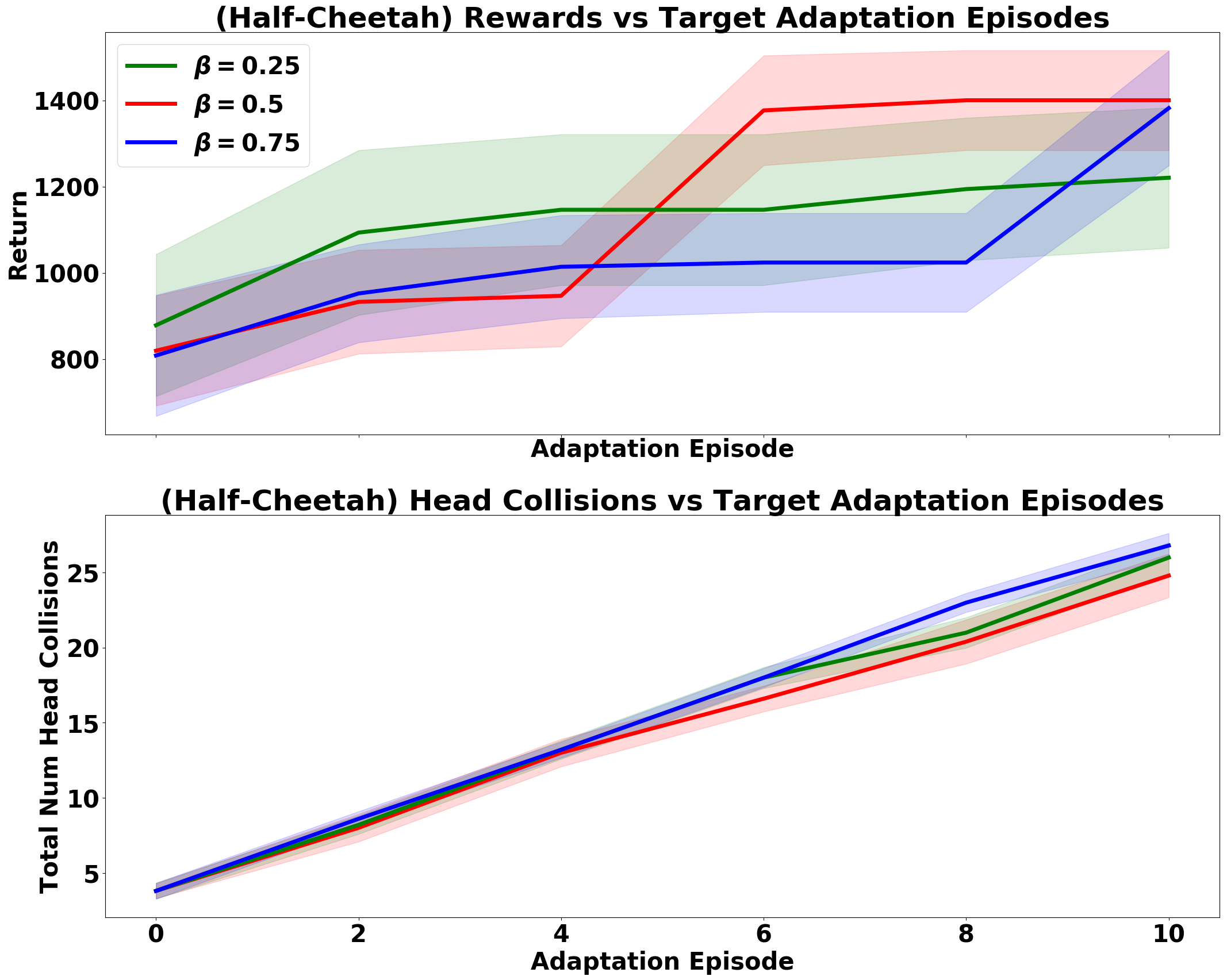}
   \label{fig:ablation}
   \caption{Evaluation of how $\beta$ affects \method\ performance in all 4 environments. A higher $\beta$ indicates less cautiousness, as the threshold for positively predicting catastrophe by each individual model in the ensemble increases, and vice versa.}
\end{figure*}
\begin{figure*}[t]
\centering
\begin{subfigure}
    \centering
    \includegraphics[width=0.475\textwidth]{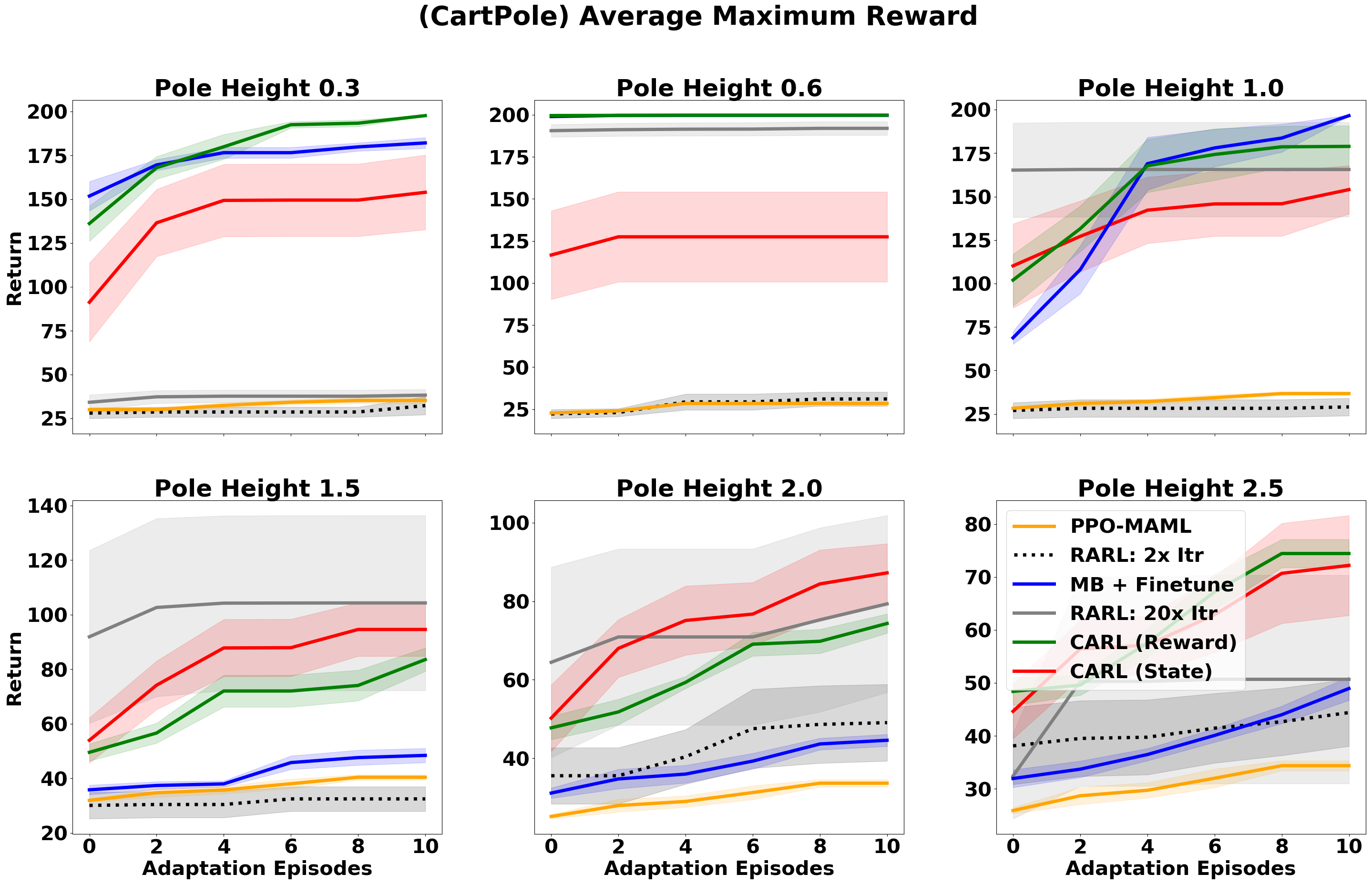}
\end{subfigure}
\begin{subfigure}
    \centering
    \includegraphics[width=0.475\textwidth]{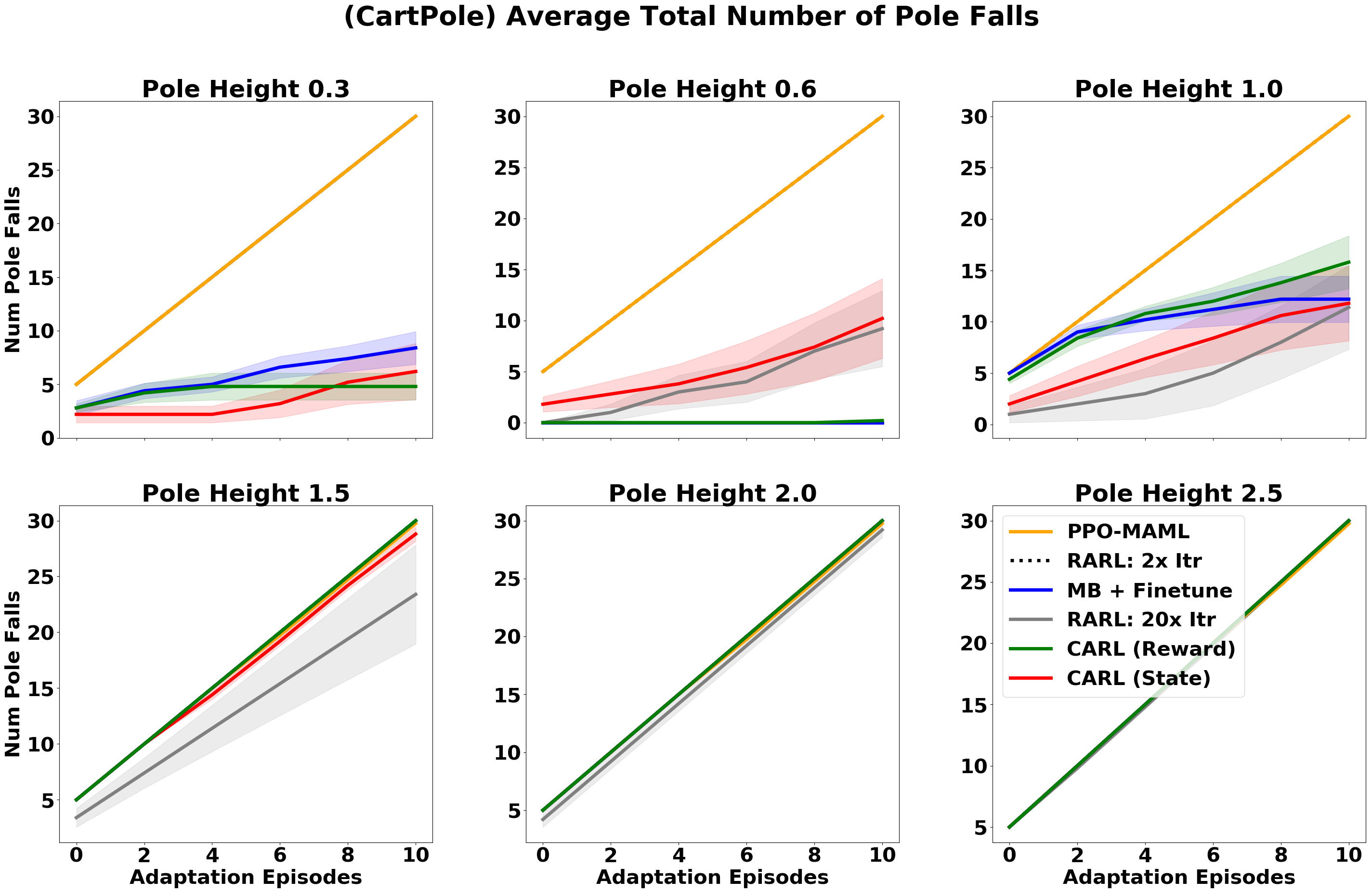}
\end{subfigure}
\begin{subfigure}
    \centering
    \includegraphics[width=0.475\textwidth]{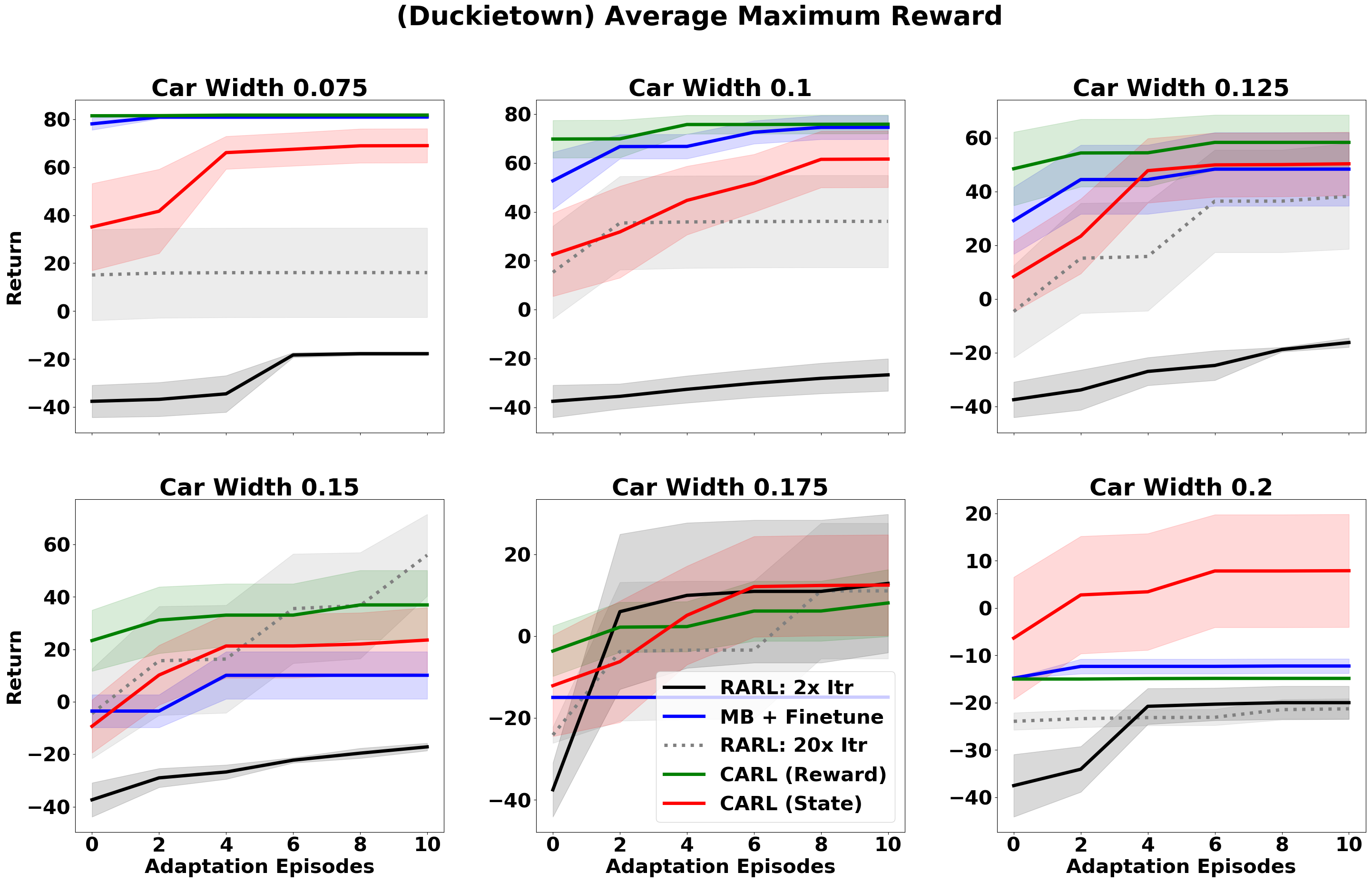}
\end{subfigure}
\begin{subfigure}
    \centering
    \includegraphics[width=0.475\textwidth]{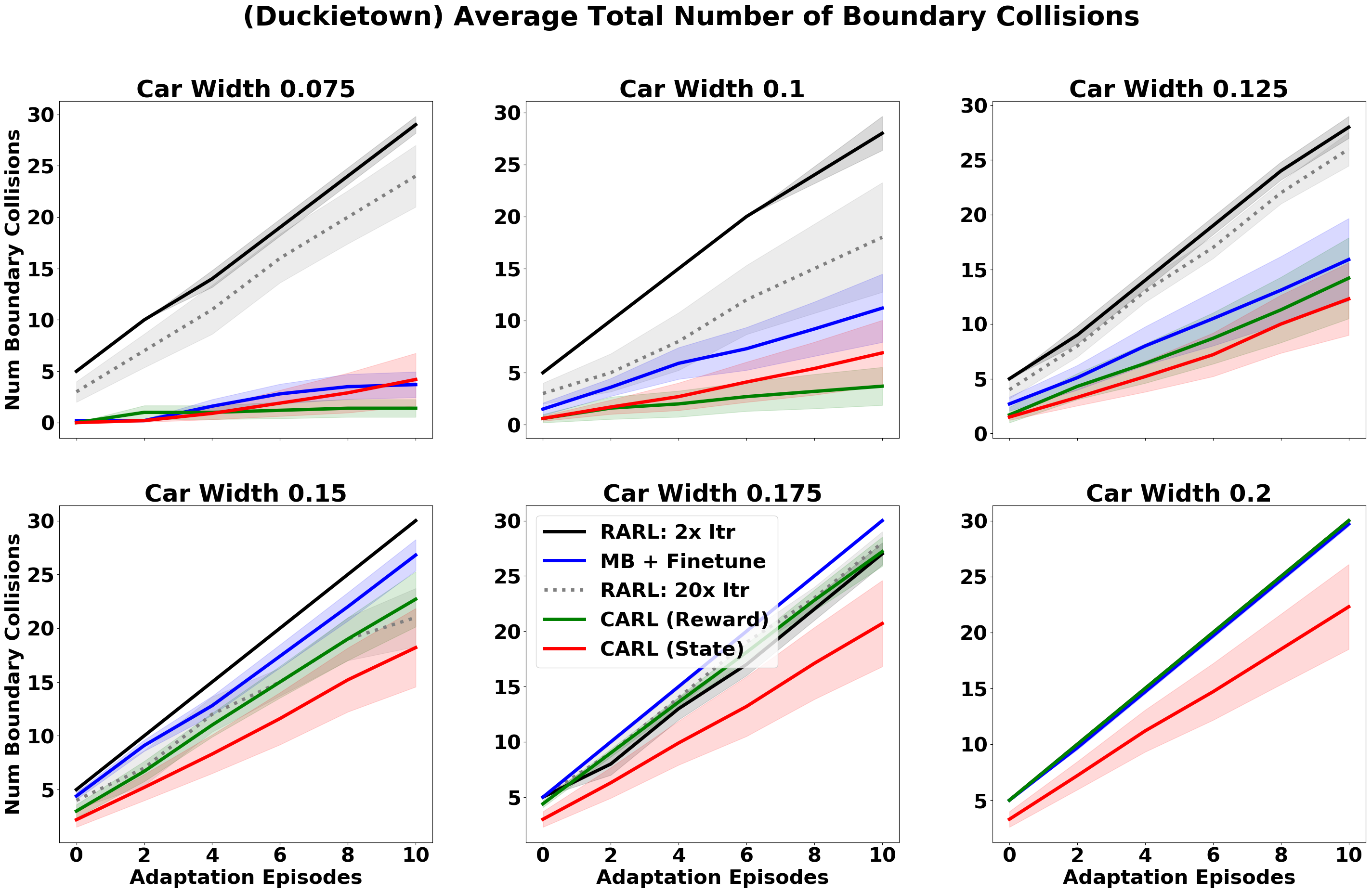}
\end{subfigure}
\begin{subfigure}
    \centering
    \includegraphics[width=0.475\textwidth]{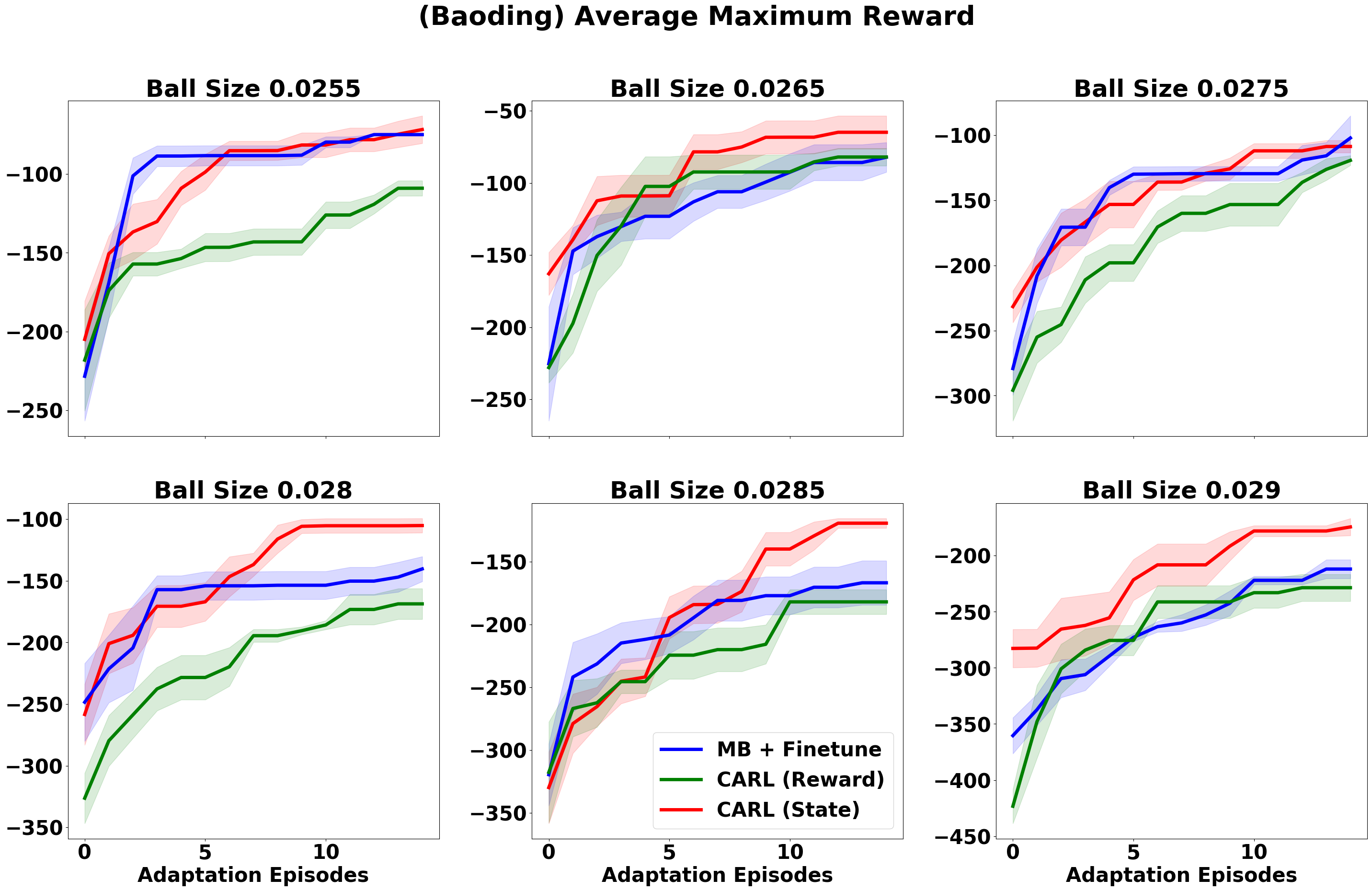}
\end{subfigure}
\begin{subfigure}
    \centering
    \includegraphics[width=0.475\textwidth]{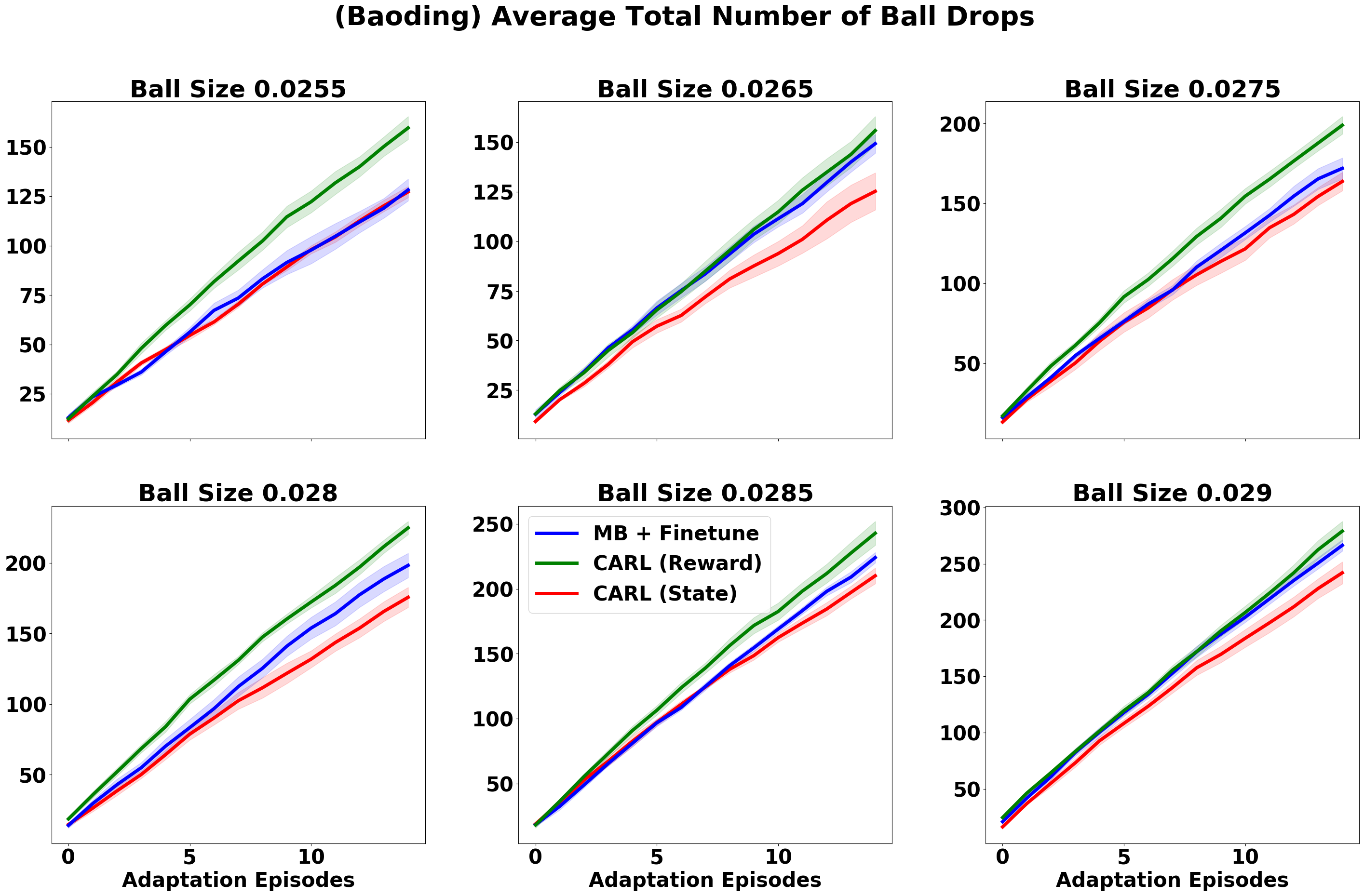}
\end{subfigure}
\label{fig:duckietown_cartpole_speed_plot}
\caption{Evaluation of the total number of catastrophes and average rewards over time at all adaptation domains in CartPole, Duckietown, and Baoding. The maximum number of catastrophes is 30 in CartPole and Duckietown, as there are 6 evaluation steps (1 before adaptation starts, 5 during) and at each evaluation step 5 evaluations are performed. The maximum catastrophe number in Baoding is 450, as there are 15 evaluation steps (1 at each adaptation iteration), with 30 evaluations performed. This cumulative total is averaged over ten initializations for each model. Each reward is calculated by taking the maximum reward seen so far at each timestep, where the value at that timestep is averaged over the  evaluations. This maximum is then averaged over ten initializations for each model. Standard errors are shown with colored bars.}
\end{figure*}